\documentclass{article} 
\usepackage{colm2024_conference}

\usepackage{booktabs}
\usepackage{graphicx}
\usepackage{enumitem}
\usepackage{wrapfig}
\usepackage{algorithm}
\usepackage{algpseudocode}
\usepackage{natbib}
\usepackage{makecell}
\usepackage{booktabs}
\usepackage{array}
\usepackage{amsmath}
\usepackage{amssymb}
\usepackage{amsfonts}
\usepackage{multirow}
\usepackage{verbatim}
\usepackage{caption}
\usepackage{longtable}
\usepackage{supertabular}
\usepackage{CJKutf8}
\usepackage[utf8]{inputenc} 
\usepackage[T1]{fontenc}
\usepackage[french,vietnamese,mongolian,greek,english]{babel}
\usepackage{pifont}

\usepackage{enumitem}
\usepackage{tablefootnote}
\usepackage{xspace}
\usepackage{textcomp}
\usepackage{makecell}
\usepackage{lscape} 
\usepackage{siunitx}
\usepackage{listings}
\usepackage{xcolor}
\usepackage{colortbl} 

\definecolor{dt}{gray}{0.7}
\usepackage{pifont}       
\usepackage{bbding}       
\usepackage{fontawesome}

\usepackage{scrextend}

\usepackage{tgpagella}
\usepackage{latexsym}
\usepackage[T1]{fontenc}
\usepackage[utf8]{inputenc}
\usepackage{microtype}
\definecolor{mydarkblue}{rgb}{0,0.08,0.45}
\definecolor{citecolor}{HTML}{0071BC}
\usepackage{url}            
\usepackage{nicefrac}       
\usepackage{changepage}
\usepackage{xargs}          
\usepackage{wrapfig,lipsum,booktabs}
\usepackage{longtable}
\usepackage{subcaption}
\usepackage{endnotes}

\providecommand{\mymodel}{DefaultModel}
\renewcommand{\mymodel}{\textbf{LexPro-1.0}}

\usepackage{pgfplots}
\usetikzlibrary{pgfplots.groupplots}
\pgfplotsset{compat=1.3}
\usepackage{tikz}
\usetikzlibrary{patterns}

\usepackage[most]{tcolorbox}

\usepackage[capitalize,noabbrev]{cleveref}
\crefname{section}{Section}{\S\S}
\Crefname{section}{Section}{\S\S}
\crefname{table}{Table}{Tables}
\crefname{figure}{Figure}{Figures}
\crefname{algorithm}{Algorithm}{}
\crefname{equation}{eq.}{}
\crefname{appendix}{Appendix}{}
\crefformat{section}{Section #2#1#3}
\usepackage{multicol}
\usepackage{tcolorbox}
\usepackage{titlesec}
\usepackage{CJK}
\usepackage{tabularx}

\titleformat*{\section}{\large\bfseries}

\usepackage{hyperref} 
\hypersetup{
    linkcolor=blue,    
}

\title{LexPro-1.0 Technical Report}

\author{
\bf Haotian Chen\thanks{haotian.chen@mail.sdu.edu.cn}, ~Yanyu Xu, ~Boyan Wang, ~Chaoyue Zhao, ~Xiaoyu Han,\\
~Fang Wang, ~Lizhen Cui, ~Yonghui Xu\thanks{xuyonghui@sdu.edu.cn}
\\
\small
\texttt{School of Software \& CFAIR \& Data Science Institute, Shandong University}
\\
}

\begin{document}

\maketitle

\begin{abstract}
In this report, we introduce our first-generation reasoning model, \mymodel, a large language model designed for the highly specialized Chinese legal domain, offering comprehensive capabilities to meet diverse realistic needs. Existing legal LLMs face two primary challenges. Firstly, their design and evaluation are predominantly driven by computer science perspectives, leading to insufficient incorporation of legal expertise and logic, which is crucial for high-precision legal applications, such as handling complex prosecutorial tasks. Secondly, these models often underperform due to a lack of comprehensive training data from the legal domain, limiting their ability to effectively address real-world legal scenarios. To address this, we first compile millions of legal documents covering over 20 types of crimes from 31 provinces in China for model training. From the extensive dataset, we further select high-quality for supervised fine-tuning, ensuring enhanced relevance and precision. The model further undergoes large-scale reinforcement learning without additional supervision, emphasizing the enhancement of its reasoning capabilities and explainability. To validate its effectiveness in complex legal applications, we also conduct human evaluations with legal experts. We develop fine-tuned models based on DeepSeek-R1-Distilled versions, available in three dense configurations: 14B, 32B, and 70B.
\end{abstract}

\section{Introduction}
In recent years, Large Language Models (LLMs) have rapidly evolved and iterated~\citep{guo2025deepseek,achiam2023gpt,yang2024qwen2}, driving increased research interest in their applications across various specialized domains. Meanwhile, LLMs have significantly influenced the workflows of legal practitioners and the advancement of the legal filed~\citep{zhou2025lawgpt}. For example, in prosecutorial work, determining convictions and sentencing often requires analyzing large volumes of legal documents. This process may involve tasks such as extracting key legal elements, conducting similar case searches, and more. By incorporating LLMs, these text-intensive tasks can be processed more efficiently, enhancing accuracy and reducing the workload of legal professionals. Recent studies indicate that GPT-4 has demonstrated the ability to pass the U.S. Judicial Exam~\citep{katz2024gpt}. Moreover, LLMs specifically trained on Chinese law have shown strong capabilities in generating legal text~\citep{dai2023laiw,li2024lexeval}. As a result, the integration of LLMs into legal proceedings has become a growing trend among legal practitioners.

Despite the significant potential of LLMs, their application in the legal domain presents several challenges due to the field's stringent requirements for expertise and precision. Firstly, the lack of high-quality, domain-specific data remains a major obstacle. Most existing legal LLMs rely on limited public datasets, restricting their ability to capture the depth and nuance of legal reasoning, ultimately leading to suboptimal performance. Secondly, hallucination remains a critical issue. As probabilistic models, LLMs can generate misleading or factually incorrect content~\citep{huang2025survey}, which is particularly concerning in the legal field. Inaccurate legal texts or flawed judicial guidance may mislead practitioners, potentially increasing their workload and leading to erroneous legal decisions. Finally, the design of evaluation metrics and tasks often fails to align with real-world legal applications. Many assessments rely on artificial benchmarks, such as multiple-choice questions~\citep{fei2023lawbench}, which do not accurately reflect the complexity of legal reasoning and practical case handling, limiting their relevance in assessing a model’s true capabilities.

Moreover, even with substantial investments in data collection and model development, prosecutors remain hesitant to adopt LLMs as a tool in their work. This reluctance stems from the lack of transparency in how LLMs generate their outputs, making it difficult to trust the results. Additionally, the generated content often suffers from poor readability and limited interpretability, further hindering its practical application in legal decision-making. Furthermore, some researchers haver leveraged existing natural language processing~(NLP) datasets to construct benchmarks for evaluating LLM performance in the Chinese legal system. However, these traditional datasets are often designed with a computer-centric perspective, focusing on isolated capabilities rather than real-world legal applications. As a result, they fail to accurately capture the practical challenges and requirements of legal professionals when utilizing LLMs in real casework.

In this report, we introduce \mymodel~(\begin{CJK*}{UTF8}{gkai}律智\end{CJK*}), a large language model specifically designed for the Chinese legal domain. To support its development, we curate and train on a comprehensive dataset comprising millions of legal documents. Beyond fundamental legal knowledge, this dataset encompasses over 20 categories of crimes from 31 provinces in China. Leveraging this dataset, we design specialized tasks tailored to different realistic legal scenarios, including legal element extraction for various crimes, named entity recognition in legal documents, legal document summarization, and similar case recommendations. During the fine-tuning stage, we adopt a two-stage supervised fine-tuning (SFT) strategy to facilitate the model’s transfer of general language capabilities to specialized legal tasks. This approach enhances the model’s ability to comprehend and apply legal knowledge, including laws, regulations, and case precedents, ensuring a more domain-aware understanding. Building upon the SFT phase, we further refine the model using large-scale reinforcement learning (RL) without explicit supervision to enhance its readability and reasoning capabilities. This stage optimizes the model’s responses by reinforcing coherence, logical consistency, and contextual alignment with legal discourse. In the inference stage, we integrate retrieval-based augmentation to improve the model’s ability to retrieve and incorporate relevant legal information dynamically.

The main characteristics of \mymodel~are as follows:

\begin{itemize}
    \item \textbf{Comprehensive Legal Knowledge Base:} \mymodel~is trained on an extensive collection of legal data, significantly surpassing existing models in scale. This dataset comprises millions of legal documents, covering various crimes from all provinces in China. The large-scale training data enhances the model’s ability to understand complex legal contexts, improving factual accuracy, reasoning capabilities, and adaptability to diverse legal scenarios. Both pre-training and post-training datasets have been carefully expanded and refined to further strengthen the model’s performance.
    \item \textbf{Efficient and Effective Model Adaptation:} we employ a multi-stage fine-tuning strategy incorporating SFT, RL, and RAG to optimize the model’s performance. SFT facilitates the acquisition and application of domain-specific legal knowledge, ensuring a strong foundation in legal reasoning. RL enhances the model’s logical coherence, readability, and interpretability, refining its ability to generate well-structured and contextually relevant responses. RAG further improves efficiency by enabling dynamic retrieval of pertinent legal information, ensuring greater factual accuracy and relevance in real-world legal applications.
    \item \textbf{Realistic Task Design and Extension Evaluation:} our task framework is designed to closely align with real-world prosecutorial workflows, ensuring practical applicability across various legal scenarios, i.e., each crime category follows specific legal rules, reflecting the diverse and complex nature of legal decision-making. Additionally, our evaluation strategy incorporates a hybrid approach combining expert assessment and automated metrics, enabling a more comprehensive evaluation of the model’s performance.
\end{itemize}

\section{Approach}
In this section, we first outline the construction of the training dataset, followed by a detailed overview of the training process.

\begin{figure*}[!t]
\centering
\includegraphics[width= 0.9\linewidth]{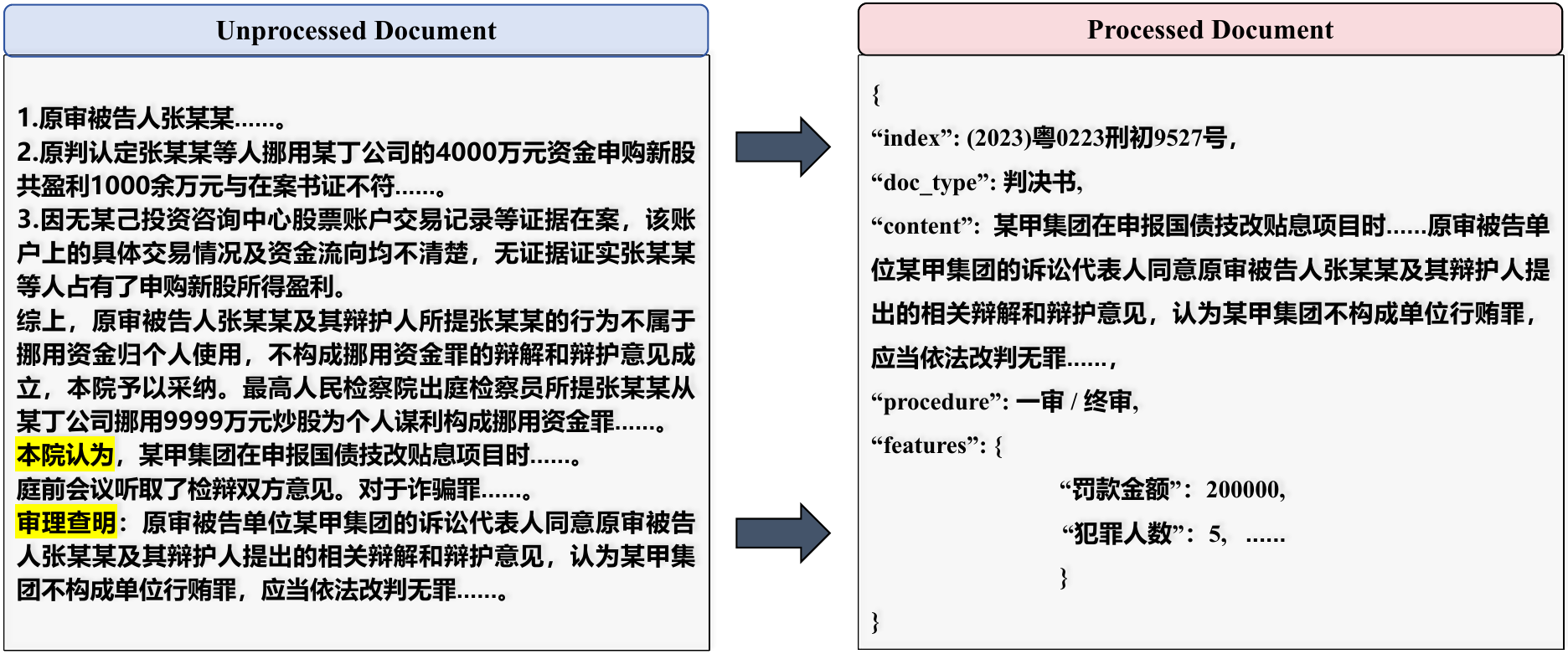}
   \caption{An example of a processed document: We extract key content from the original (unprocessed) document and structure the essential information into JSON format for subsequent training.
}
\label{fig:json_data_example}
\end{figure*}

\subsection{Training Data}
\label{sec:training_data}
We collect and organize millions of legal documents, including over 20 crime types from over 31 provinces from mainland China. 

Taking \textit{Crime of Intentional Injury}~(\begin{CJK*}{UTF8}{gkai}故意伤害罪\end{CJK*}) as an example, we construct the dataset by extracting key sections from original legal documents. Specifically, we retain only the critical segments that form the substantive content of the judgment, such as \textit{"This court found out," "This court believes," "The judgment is as follows," "The judgment result,"} and \textit{"The trial found out"}~(\begin{CJK*}{UTF8}{gkai}"本院查明", "本院认为", "判决如下", "裁判结果", "审理查明"\end{CJK*}). This selection ensures that the dataset captures essential legal reasoning and judicial conclusions while eliminating extraneous information. Additionally, necessary metadata, including \textit{"index," "doc\_type," "procedure,"} and \textit{"features,"} is incorporated to structure the data in JSON format for subsequent training. An illustration of the processed JSON data is provided in Figure~\ref{fig:json_data_example}. 

Due to the complexity of legal documents, certain types are more suitable for training than others. For instance, a criminal ruling~(\begin{CJK*}{UTF8}{gkai}裁定书\end{CJK*}) primarily addresses procedural matters and may be subject to further review or appeal, making it unsuitable as a final judgment. In contrast, a criminal judgment~(\begin{CJK*}{UTF8}{gkai}判决书\end{CJK*}) generally carries finality, particularly after a second-instance trial, where it becomes legally binding and cannot be retried. Therefore, we predominantly select judgments for our training dataset while filtering out other document types to ensure legal consistency and reliability. Due to the evolving nature of laws, where newly enacted laws do not retroactively apply to past cases, older legal documents may have limited reference value for training. To ensure relevance and alignment with current legal standards, we selectively include only documents from the year 2020 onward in our main dataset.

\subsection{Method Overview}  
Our language model adaptation process consists of two key stages: post-training and inference. In the post-training stage, we meticulously curate high-quality legal data through advanced filtering and scoring mechanisms to construct a refined dataset for SFT. Following the initial SFT, we further augment the high-quality data and conduct a secondary SFT to enhance the model’s understanding of legal knowledge and its ability to handle complex legal tasks with greater precision. Subsequently, we apply large-scale RL without explicit supervision to improve the model’s readability and reasoning capabilities, ensuring coherent and contextually accurate responses. During the inference stage, we integrate RAG to dynamically retrieve relevant legal information, enhancing factual accuracy and ensuring that the model’s outputs remain grounded in reliable legal sources.

\subsection{Post-training}
In the post-training stage, our goal is to adapt a general language model to effectively handle legal tasks by rapidly acquiring domain-specific knowledge. To achieve this, we employ a two-pronged approach: SFT to systematically enhance the model’s understanding of legal concepts and RL to refine its reasoning quality for more accurate and reliable legal analysis.

\subsubsection{Supervised Fine-Tuning}
SFT is essential for adapting large language models to the legal domain, which requires a high degree of specialization. General pre-trained models may capture broad linguistic patterns but often lack the domain-specific legal reasoning and terminology necessary for accurate responses. By fine-tuning on high-quality legal corpora, we enable the model to better understand statutory language, case law, and legal doctrines, thereby improving its ability to generate precise and contextually relevant legal texts. This process helps bridge the gap between general language understanding and the intricate demands of legal applications, ensuring the model aligns more closely with professional legal reasoning.

\textbf{Fundamental Legal Knowledge} We curated a collection of essential legal texts and their corresponding judicial interpretations to enhance the model’s understanding of core legal principles. These include the \textit{ "Criminal Law of the People's Republic of China", "the Civil Code of the People's Republic of China", "the Constitution of the People's Republic of China", and "the Criminal Procedure Law of the People's Republic of China"}~(\begin{CJK*}{UTF8}{gkai}中华人民共和国刑法, 中华人民共和国民法典, 中华人民共和国宪法, 中华人民共和国刑事诉讼法\end{CJK*}), among others. We optimize the model using a supervised fine-tuning objective based on cross-entropy loss, which is defined as:  
\begin{equation}
\label{eq:loss_sft}
    \mathcal{L}_{SFT}(\theta) = - \sum_{(x, y^*) \in D} \sum_{t=1}^{T} \log P_\theta(y^*_t \mid x, y^*_{<t})
\end{equation}

where \( x = (x_1, x_2, ..., x_N) \) represents the input token sequence, and \( y^* = (y^*_1, y^*_2, ..., y^*_T) \) is the corresponding ground truth target sequence. The dataset \( D \) consists of multiple input-output pairs \( (x, y^*) \), and \( P_\theta(y^*_t \mid x, y^*_{<t}) \) denotes the probability distribution predicted by the model for the \( t \)-th token, conditioned on the input sequence and previously generated tokens. The sequence length \( T \) corresponds to the number of tokens in the target output. The objective \( \mathcal{L}_{SFT}(\theta) \) is minimized to refine model predictions during supervised fine-tuning.

\textbf{Specialized Legal Knowledge} Building upon fundamental legal knowledge, the next step is to acquire more specialized legal expertise. One critical aspect is legal element extraction, a fundamental task in the legal domain that structures legal texts, enabling more precise and systematic legal analysis. This process is essential for various legal intelligence applications, including similar case retrieval and case summary generation. To develop high-quality training data, we first employ a combination of rule-based extraction and manual verification to annotate a subset of the dataset. For instance, in cases involving the Crime of Intentional Injur, we extract more than 79 legal elements, such as "Control," "Detention," "Fixed-term imprisonment," "Life imprisonment," "Death penalty," and "Probation". These refined annotations are then further processed to create high-quality datasets for SFT, ensuring the model's ability to recognize and apply legal elements effectively.

\textbf{Knowledge Augmentation} Existing research indicates that LLM-based data generation methods have significant potential for constructing high-quality training datasets~\citep{li2025neuro}. To further improve the model’s comprehension of legal knowledge, we utilize LLMs to generate additional training data by providing structured legal inputs. As shown in Table~\ref{tab:template_generation}, we design tailored prompts that guide the model to produce multiple question-answer (QA) pairs based on the given legal context. This approach enriches the training corpus, reinforcing the model’s ability to interpret and apply legal concepts more effectively.

\begin{table}[t]
    \centering
    \small
    \renewcommand{\arraystretch}{1.2}
    \begin{tabularx}{\textwidth}{|X|}
    \hline
    \textbf{Prompt Template of Data Generation} \\
    \hline
    A conversation between User and Assistant. The user asks a question, and the Assistant solves it. \\
    I am fine-tuning a large legal model and need to generate some question-answer pairs (QA) based on legal text as data enhancement. \\
    The following is a piece of legal text:~\textcolor{red}{\{prompt\}} \\ 
    Please generate \textcolor{red}{\{num\_qa\}} high-quality question-answer pairs (QA) based on this text. Both the questions and answers are required to be based on the text content, and the questions should cover the key legal concepts, clauses, or principles in the text. \\ 
    
    \textbf{The sample format is as follows:} \\
    \{ \\
    \quad "input": "What is Article 96 of the Civil Code?",  \\
    \quad "output": "The legal persons of institutions......" \\
    \} \\

    \textbf{Please ensure that the generated QA pairs meet the following requirements:} \\
    1. The questions are clear, and the answers are accurate and directly derived from the text. \\
    2. Question types may include definitions, interpretation of terms, scope of application, legal liability, etc. \\
    3. The answer should be as concise as possible and avoid redundant information. \\

    \textbf{Please return the QA pair in the following format:} \\
    \verb|[ { "input": "Question 1", "output": "Answer 1" }, ... ]| \\
    \hline
    \end{tabularx}
    \caption{A template for data generation, where \textcolor{red}{prompt} represents the input legal knowledge data, and the model generates corresponding outputs based on this input. The parameter \textcolor{red}{num\_qa} specifies the number of generated data samples.}
    \label{tab:template_generation}
\end{table}

\subsubsection{Reinforcement Learning}  
Labeling millions of legal documents is an impractical and time-intensive task. Recent advancements have shown that reinforcement learning (RL) can be highly effective for reasoning tasks, even in the absence of supervised data, and can achieve self-improvement through purely reinforcement-based training~\citep{guo2025deepseek,shao2024deepseekmath}. Inspired by these findings, we leverage large-scale RL to enhance the model’s readability and explainability, making its outputs more accessible and interpretable for legal practitioners, thereby lowering the barrier to adoption in real-world legal applications.

To reduce the computational cost of RL training, we adopt Group Relative Policy Optimization (GRPO)~\citep{shao2024deepseekmath}, which removes the need for maintaining a separate critic model. Since the critic model is usually the same size as the policy model, a large policy model would result in significantly higher computational costs. Additionally, it computes the advantage function using relative rewards within the group, eliminating reliance on traditional value functions (e.g., the Critic model). In contrast, conventional reward models may introduce estimation errors that impact decision-making, which can have serious consequences in the legal domain. For a given query $q$, GRPO generates a set of $G$ candidate outputs $\{o_1, \dots, o_G\}$ using the old policy $\pi_{\theta_{\text{old}}}$. Instead of relying on a value function, it evaluates these outputs relative to each other and updates the new policy $\pi_{\theta}$ by optimizing the following objective:

\begin{equation}
\label{eq:gpro}
    \left\{
\begin{aligned}
J_{GRPO}(\theta) &= \mathbb{E}_{q \sim P(Q), \{o_i\}_{i=1}^{G} \sim \pi_{\theta_{\text{old}}} (O|q)} 
\left[
\frac{1}{G} \sum_{i=1}^{G} \left( \min \left( \frac{\pi_{\theta}(o_i|q)}{\pi_{\theta_{\text{old}}}(o_i|q)} A_i, 1 - \epsilon, 1 + \epsilon \right) A_i \right) - \beta D_{KL}(\pi_{\theta} || \pi_{\text{ref}})
\right] \\[10pt]
D_{KL}(\pi_{\theta} || \pi_{\text{ref}}) &= \sum_{o} \pi_{\theta}(o|q) \log \frac{\pi_{\theta}(o|q)}{\pi_{\text{ref}}(o|q)} \\[10pt]
A_i &= \frac{r_i - \text{mean}(\{r_1, r_2, \dots, r_G\})}{\text{std}(\{r_1, r_2, \dots, r_G\})}
\end{aligned}
\right.
\end{equation}

where $\epsilon$ and $\beta$ are hyperparameters that control the update constraints and regularization strength, respectively. The advantage $A_i$ is computed based on a group of rewards $\{r_1, r_2, \dots, r_G\}$ assigned to the sampled outputs, ensuring that each output is evaluated in relation to its peers within the group. Moreover, the KL divergence term $D_{KL}(\pi_{\theta} || \pi_{\text{ref}})$ serves as a regularization factor, preventing the updated policy $\pi_{\theta}$ from deviating too far from the reference policy $\pi_{\text{ref}}$. This helps stabilize training and avoids overly aggressive updates.

\begin{table}[t]
    \centering
    \small
    \renewcommand{\arraystretch}{1.2}
    \begin{tabularx}{\textwidth}{|X|}
    \hline
    \textbf{Summary Template} \\
    \hline
    When the party applies for supervision, their statement is: \textcolor{red}{[applicant]} believes that
    \textcolor{red}{[court]} People's Court's trial of \textcolor{red}{[case information]} has 
    legal violations and requests the court to supervise it. This case has now concluded the review of the supervision. 
    \newline
    (The expression of the court’s internal review findings is:) This court has reviewed the trial activities of 
    \textcolor{red}{[court]} People's Court in the case of \textcolor{red}{[case information]}. 
    The review has been completed. Now, it is clarified: 
    \newline
    (Provide detailed and clear explanations of the specific judicial actions taken by the People's Court in the case.)
    (If necessary, include a conclusion regarding the review and clarify the reasoning and basis for the trial activities.)
    \newline
    In summary, (list which judicial actions of the People's Court complied with or violated the law, based on the specific legal provisions),
    according to Article \textcolor{red}{[law article]} of the Administrative Procedure Law of the People's Republic of China,
    we hereby issue this supervision decision: (write the specific content of the decision). \\
    \hline
    \end{tabularx}
    \caption{A structured template for legal document summaries, where \textcolor{red}{[xxx]} represents placeholders for specific legal content to be filled in based on the document's details.}
    \label{tab:summary_template}
\end{table}

\textbf{Reward Modeling} The core aspect of RL training lies in the design of the reward function. Given the scarcity of labeled data, we implement a rule-based reward system, which primarily consists of two types of rewards:

\begin{itemize}
    \item \textbf{Format Rewards:} Legal documents are governed by strict formatting regulations to ensure consistency and clarity. For example, criminal record forms require a structured layout documenting key details such as the defendant, criminal facts, and evidence, facilitating systematic analysis. Similarly, case summaries must adhere to predefined formats to maintain coherence and usability, as illustrated in Table~\ref{tab:summary_template}. To enforce proper structuring across different tasks, we incorporate a format reward mechanism that encourages the model to generate outputs in the appropriate format for each specific legal task.
    \item \textbf{Process Rewards:} To improve readability and interpretability, we encourage the model to generate intermediate reasoning steps before producing the final output. This structured reasoning process enhances transparency and helps users better understand how conclusions are reached.  For example, when extracting the "criminal amount" from legal documents, the model is guided to follow a step-by-step approach:  
    \begin{enumerate}
        \item Identify the total number of robbery incidents.  
        \item Extract the details of each incident, including the specific items stolen.  
        \item Determine the value of each stolen item.  
        \item Summarize the total amount involved in the crime.  
    \end{enumerate}  
    By breaking down the reasoning process in this manner, the model improves both the accuracy of its extractions and the comprehensibility of its outputs, making legal decision support more reliable

\end{itemize}

Ultimately, a diverse data distribution enables us to train a model that not only excels in legal reasoning but also prioritizes helpfulness and harmlessness. This ensures that legal practitioners can effectively utilize the model to enhance their workflow and decision-making processes in real-world legal applications.

\subsection{Inference}
In the inference stage, we address the challenges of inefficiency and inaccuracy in standalone model outputs caused by the length and complexity of legal documents and instructions.

\subsubsection{Retrieval-Based Augmentation}
Given that different crimes require the extraction of distinct legal elements, and these elements can be numerous and embedded within lengthy legal documents, directly inputting the entire text into an LLM is inefficient and computationally expensive. Additionally, excessive input length can lead to context fragmentation and reduced processing accuracy. To address this, we integrate retrieval-based augmentation to enhance the model’s efficiency and precision. Instead of feeding the full document into the LLM, RAG first segments the legal text, retrieves the most relevant legal elements for the given task, and then provides a refined context to the model. 

Given a legal document \( t \), we first segment it into meaningful chunks \( T = \{t_1, t_2, ..., t_n\} \). Each segment \( t_i \) is encoded into an embedding vector \( \mathbf{E}(t_i) \), and we compute its similarity with predefined legal elements \( E = \{e_1, e_2, ..., e_m\} \) to retrieve the most relevant elements \( D_i \), using:
\begin{equation}
D_i = \arg\max_{e \in E} \frac{\mathbf{E}(t_i) \cdot \mathbf{E}(e)}{\|\mathbf{E}(t_i)\| \|\mathbf{E}(e)\|}
\end{equation}
where \( \mathbf{E}(t_i) \) and \( \mathbf{E}(e) \) are embeddings of segment \( t_i \) and legal element \( e \), respectively. The retrieved legal elements \( D = \{D_1, ..., D_n\} \) are combined with the original document \( t \) and fed into an LLM to generate the final structured output:
\begin{equation}
P(y | t, D) = \prod_{j=1}^{T} P(y_j | t, D, y_{<j})
\end{equation}
where \( y \) represents the extracted legal elements.

\section{Experiments}
In this section, we evaluate the model's performance on two major datasets: a fundamental legal knowledge dataset and our curated high-quality legal document dataset.

\textbf{Experimental Setup}  We fine-tune our model using the DeepSeek-R1-Distilled versions with parameter sizes of 14B, 32B, and 70B~\footnote{https://huggingface.co/unsloth}. The experiments are implemented in PyTorch and conducted on NVIDIA L40 GPUs. We set the batch size to 2, with gradient accumulation steps of 4 to accommodate large-scale training. The learning rate is configured as $2e-4$, and we utilize the AdamW\_8bit optimizer to improve memory efficiency and training stability.

We configure the maximum sequence length to 2048 and set the maximum generation length to 512. For text generation tasks, we employ ROUGE-1 (R1), ROUGE-2 (R2), ROUGE-L (RL)~\citep{lin2004rouge}, and BERTScore~\citep{bert-score} as evaluation metrics to assess the quality and relevance of the generated text.  

\subsection{Basic Legal Knowledge}

\begin{figure*}[!t]
\centering
\includegraphics[width=\linewidth]{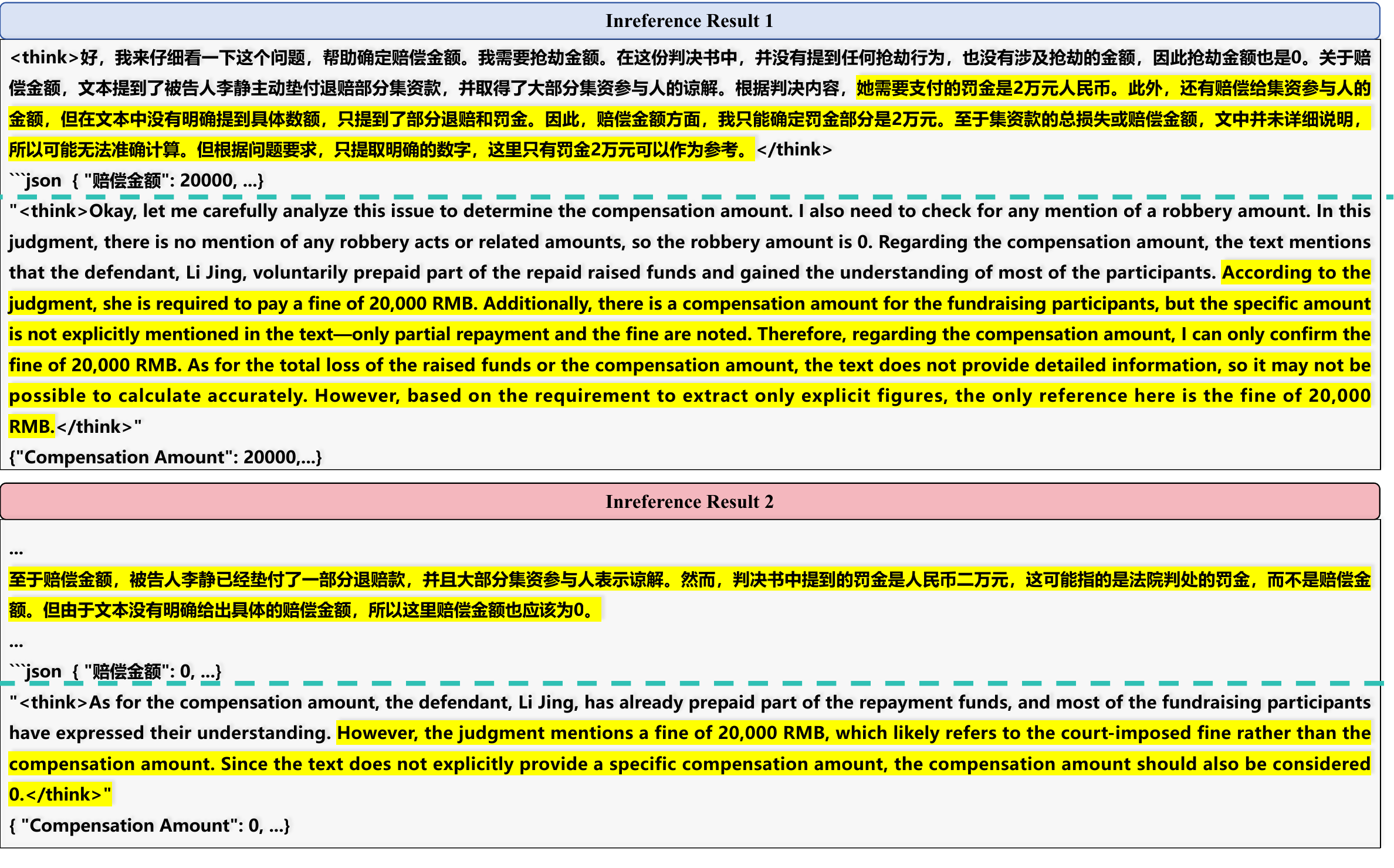}
\caption{We input the same example twice, and the model produced different results each time. This inconsistency arises due to the model's incomplete legal knowledge, making it unable to clearly distinguish between fines and compensation, leading to unstable outputs.}
\label{fig:legal_motivation}
\end{figure*}

\begin{table*}[!t]
    \centering
\caption{An example of data augmentation: we construct more complex QA pairs based on basic QA to enhance understanding capabilities.}
    \label{tab:law_qa_aug_examples}
    \renewcommand{\arraystretch}{1.2}  
    \resizebox{1.0\textwidth}{!}{  
        \begin{tabular}{>{\centering\arraybackslash}p{0.5\textwidth}|>{\centering\arraybackslash}p{0.5\textwidth}}  
            \toprule
            Question & Answer \\
               \midrule
            \multicolumn{2}{c}{\emph{Dataset (Raw) }}\\
            \midrule
    \begin{CJK*}{UTF8}{gkai}刑法第一条的内容是什么？\end{CJK*} & \begin{CJK*}{UTF8}{gkai}【立法宗旨】为了惩罚犯罪，保护人民，根据宪法，结合我国同犯罪作斗争的具体经验及实际情况，制定本法。\end{CJK*} \\
What is the content of Article 1 of the Criminal Law? & [Legislative Purpose] This Law is formulated in order to punish crimes, protect the people, and, based on the Constitution, in light of China's specific experience and actual conditions in combating crimes. \\
            \midrule
            \multicolumn{2}{c}{\emph{Dataset (Augmented) }}\\
            \midrule
    \begin{CJK*}{UTF8}{gkai}刑法第一条中提到的刑法制定的主要目的是什么？\end{CJK*} & \begin{CJK*}{UTF8}{gkai}为了惩罚犯罪，保护人民。\end{CJK*} \\
What is the main purpose of the formulation of criminal law mentioned in Article 1 of the Criminal Law? &To punish crime and protect the people. \\
    \begin{CJK*}{UTF8}{gkai}根据刑法第一条，刑法的制定依据是什么？\end{CJK*} & \begin{CJK*}{UTF8}{gkai}根据宪法，结合我国同犯罪作斗争的具体经验及实际情况。\end{CJK*} \\
According to Article 1 of the Criminal Law, what is the basis for the formulation of the Criminal Law? &According to the Constitution, combined with China's specific experience and actual situation in fighting crime. \\
    \begin{CJK*}{UTF8}{gkai}刑法第一条中体现了哪些原则？\end{CJK*} & \begin{CJK*}{UTF8}{gkai}立法宗旨是惩罚犯罪、保护人民，并且结合了我国同犯罪作斗争的具体经验和实际情况。\end{CJK*} \\
The legislative purpose is to punish crimes, protect the people, and is based on the specific experience and actual situation of China's fight against crime. &The legislative purpose is to punish crimes, protect the people, and is based on the specific experience and actual situation of China's fight against crime. \\
    \begin{CJK*}{UTF8}{gkai}根据刑法第一条，可以总结出该法律的核心理念是什么？\end{CJK*} & \begin{CJK*}{UTF8}{gkai}立通过立法惩治犯罪，保障人民的利益和社会秩序。\end{CJK*} \\
According to Article 1 of the Criminal Law, what is the core concept of this law? &Punish crimes through legislation to safeguard the interests of the people and social order. \\
            $\cdots$ & $\cdots$ \\
            \bottomrule
        \end{tabular}
    }
\end{table*}

We initially conducted legal element extraction tasks using the DeepSeek-R1-Distilled models and observed that their legal comprehension was insufficient. To further investigate, we tested additional examples with the DeepSeek-R1-671B model and encountered similar issues. As shown in Figure~\ref{fig:legal_motivation}, the input instruction is "Extract the compensation amount from the input text." We perform inference twice and obtain two distinct results. In the first inference result, the model identifies only the fine of 20,000 RMB as the compensation amount, potentially misinterpreting it as the total loss of the raised funds or the actual compensation amount. In contrast, in the second inference result, the model states that the judgment mentions a fine of 20,000 RMB, which likely refers to a court-imposed fine rather than the compensation amount. Since the text does not explicitly specify a distinct compensation amount, the correct conclusion should be that the compensation amount is 0. This inconsistency in outputs suggests that the model struggles with fully understanding complex and extensive legal knowledge. Therefore, as a fundamental step, we first need to fine-tune the model to enhance its comprehension of essential legal concepts.

The most straightforward approach is to construct basic QA pairs directly from legal texts. For example, a question like "What is Article 1 of the Criminal Law?" can be paired with the corresponding legal provision as the answer. However, beyond such simple QA pairs, we aim to generate more diverse and complex questions that probe deeper into the content, thereby enhancing the model’s understanding of the law. To achieve this, we leverage a 70B model to generate additional QA pairs (c.f. Table~\ref{tab:template_generation}). As shown in Table~\ref{tab:law_qa_aug_examples}, the basic QA pairs can be expanded into multiple augmented versions using LLMs. This augmented dataset is then used to fine-tune our model, significantly improving its comprehension of legal knowledge.

\begin{figure*}[!t]
\centering
\includegraphics[height=18cm]{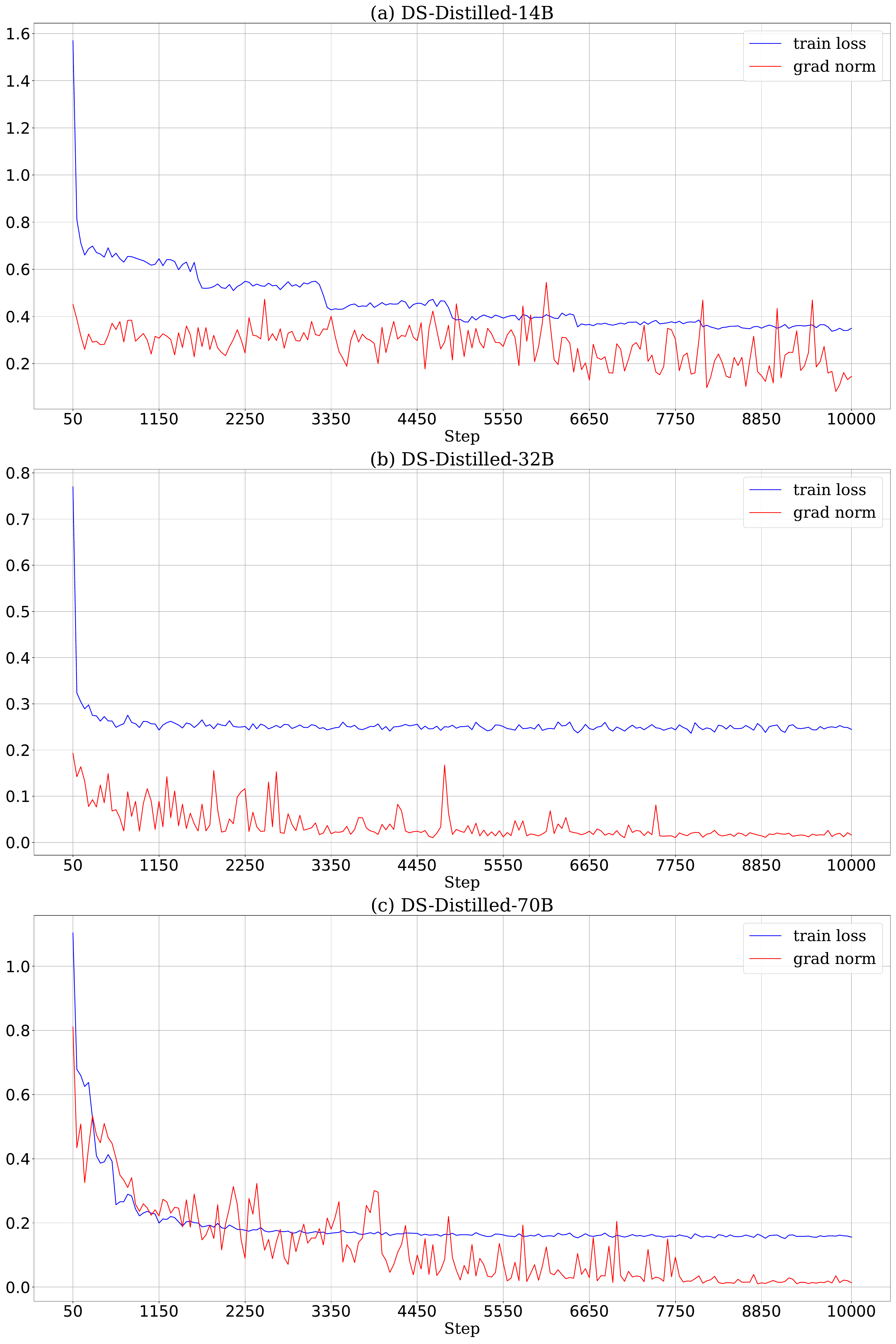}
   \caption{The average training progress of 7B, 14B, and 30B models during the SFT process, reporting training loss and gradient norm on the training set.
}
\label{fig:fine_tune_laws_loss}
\end{figure*}

We  report the performance on four basic laws: \textit{ "Criminal Law of the People's Republic of China", "the Civil Code of the People's Republic of China", "the Constitution of the People's Republic of China", and "the Criminal Procedure Law of the People's Republic of China"}. The training status during the SFT process of 14B, 32B, and 70B model is shown in Figure~\ref{fig:fine_tune_laws_loss}. We report the train loss, i.e., token cross entropy and the gradient norm.  As we can see in the figure, after 1150 steps, the loss starts to converge and the gradient norm fluctuations within a certain range. However, we evaluate the accuracy of the test dataset in 5000 steps, the result are not very good not stable, so we further fine-tune 10000 steps to ensure more stable output.

We report the performance of all models in Table~\ref{tab:basic_laws}. The ROUGE metric evaluates text similarity by calculating n-gram overlap and the longest common subsequence between the automatically generated text and the reference text. Specifically, ROUGE-1 and ROUGE-2 primarily measure word- and phrase-level matching, while ROUGE-L focuses on word order and sentence structure. Consequently, the ROUGE metric imposes certain formatting requirements; factors such as word segmentation, punctuation, capitalization, and space processing can significantly impact the final calculation results. We find that the original model outputs lack standardized formatting, leading to the lowest ROUGE scores. To further assess the quality of generated text, we also compute the semantic similarity score using BERTScore. As shown in Table~\ref{tab:basic_laws}, after supervised fine-tuning (SFT), all models exhibit substantial improvements, with performance gains exceeding 30\%. Notably, the most significant improvement is observed in ROUGE scores, indicating that the generated outputs have become more standardized. It is worth noting that DS-Distilled-14B and DS-Distilled-32B are distilled versions of the Qwen model~\citep{yang2024qwen2}, which demonstrates strong Chinese language capabilities. Moreover, scaling effects are evident, as the 32B model outperforms the 14B model. However, DS-Distilled-70B is distilled from Llama-3.3-70B-Instruct~\citep{dubey2024llama}, which explicitly states limited support for Chinese. As a result, despite having significantly more parameters, the 70B model exhibits weaker Chinese proficiency compared to Qwen 14B and 32B.

\begin{table*}[!ht]
    \centering
    \caption{Performance of basic legal knowledge.}
    \label{tab:basic_laws}
    \renewcommand{\arraystretch}{1.2}  
    \resizebox{\textwidth}{!}{  
        \begin{tabular}{l|ccc|ccc|ccc|ccc}
            \toprule
            \multirow{2}{*}{\textbf{Model}} & \multicolumn{3}{c|}{\textbf{R1}} & \multicolumn{3}{c|}{\textbf{R2}} & \multicolumn{3}{c|}{\textbf{R3}} & \multicolumn{3}{c}{\textbf{BERT Score}} \\
  
            \cmidrule(lr){2-4} \cmidrule(lr){5-7} \cmidrule(lr){8-10} \cmidrule(lr){11-13}
            & Recall & Precision & F1  & Recall & Precision & F1 & Recall & Precision & F1 & Recall & Precision & F1  \\
          \midrule
            \multicolumn{13}{c}{\emph{Criminal Law of the People's Republic of China}}\\
            \midrule
            
            DS-Distilled-14B       & 0.8  & 0.3  & 0.4  & 0.1  & 0.1  & 0.1  & 0.8  & 0.3  & 0.4  & 65.1 & 71.3 & 67.9 \\
DS-Distilled-32B        & 2.1  & 1.6  & 1.7  & 0.5  & 0.4  & 0.5  & 2.1  & 1.4  & 1.5  & 67.6 & 74.1 & 70.5 \\
DS-Distilled-70B       & 0.0  & 0.0  & 0.0  & 0.0  & 0.0  & 0.0  & 0.0  & 0.0  & 0.0  & 62.2 & 64.2 & 63.0 \\
\rowcolor{blue!8}DS-Distilled-14B (SFT) & 71.8 & 75.3 & 72.9 & 31.4 & 33.6 & 32.1 & 71.8 & 75.3 & 72.9 & 97.9 & 96.0 & 96.9 \\
\rowcolor{blue!8}DS-Distilled-32B (SFT) & 81.3 & 85.9 & 82.8 & 35.5 & 37.6 & 36.2 & 81.3 & 85.9 & 82.8 & 99.1 & 97.6 & 98.3 \\
\rowcolor{blue!8}DS-Distilled-70B (SFT) & 68.9 & 67.7 & 66.9 & 21.8 & 21.5 & 21.0 & 68.8 & 67.7 & 66.9 & 94.3 & 93.6 & 93.3  \\
            \midrule
            \multicolumn{13}{c}{\emph{Civil Code of the People's Republic of China}}\\
            \midrule
            DS-Distilled-14B       & 0.8  & 0.3  & 0.4  & 0.1  & 0.0  & 0.0  & 0.8  & 0.2  & 0.3  & 55.3 & 63.0 & 58.8 \\
DS-Distilled-32B        & 1.4  & 0.9  & 1.1  & 0.6  & 0.3  & 0.4  & 1.4  & 0.7  & 0.9  & 58.6 & 66.7 & 62.3 \\
DS-Distilled-70B       & 0.0  & 0.0  & 0.0  & 0.0  & 0.0  & 0.0  & 0.0  & 0.0  & 0.0  & 52.6 & 58.1 & 55.2 \\
\rowcolor{blue!8}DS-Distilled-14B (SFT) & 86.6 & 88.2 & 87.0 & 40.2 & 40.6 & 40.3 & 86.5 & 88.1 & 87.0 & 97.8 & 97.1 & 97.4 \\
\rowcolor{blue!8}DS-Distilled-32B (SFT) & 88.7 & 91.5 & 89.5 & 40.6 & 41.5 & 40.8 & 88.7 & 91.5 & 89.5 & 99.2 & 97.2 & 98.2 \\
\rowcolor{blue!8}DS-Distilled-70B (SFT) & 50.4 & 44.5 & 45.7 & 23.2 & 20.6 & 21.1 & 50.4 & 44.5 & 45.7 & 81.2 & 81.6 & 81.2 \\
            \midrule
            \multicolumn{13}{c}{\emph{Constitution of the People's Republic of China}}\\
            \midrule
 DS-Distilled-14B       & 1.3  & 0.8  & 0.9  & 0.3  & 0.2  & 0.2  & 1.3  & 0.6  & 0.8  & 61.9 & 67.1 & 64.3 \\
DS-Distilled-32B        & 3.9  & 2.8  & 3.3  & 1.0  & 0.6  & 0.7  & 3.9  & 1.9  & 2.5  & 65.0 & 70.8 & 67.6 \\
DS-Distilled-70B       & 0.0  & 0.0  & 0.0  & 0.0  & 0.0  & 0.0  & 0.0  & 0.0  & 0.0  & 55.8 & 59.8 & 57.6 \\
\rowcolor{blue!8}DS-Distilled-14B (SFT) & 88.3 & 92.9 & 89.6 & 55.5 & 57.5 & 56.0 & 88.3 & 92.9 & 89.6 & 98.0 & 97.9 & 97.9 \\
\rowcolor{blue!8}DS-Distilled-32B (SFT) & 91.9 & 91.7 & 91.3 & 62.7 & 62.0 & 61.7 & 91.9 & 91.7 & 91.3 & 99.4 & 98.7 & 99.0 \\
\rowcolor{blue!8}DS-Distilled-70B (SFT) & 59.0 & 57.9 & 57.9 & 32.5 & 32.8 & 32.2 & 59.0 & 57.9 & 57.9 & 89.5 & 90.6 & 90.0 \\
            \midrule
            \multicolumn{13}{c}{\emph{Criminal Procedure Law of the People's Republic of China}}\\
            \midrule
DS-Distilled-14B       & 0.0  & 0.0  & 0.0  & 0.0  & 0.0  & 0.0  & 0.0  & 0.0  & 0.0  & 57.3 & 62.3 & 59.6 \\
DS-Distilled-32B        & 0.6  & 0.5  & 0.5  & 0.4  & 0.3  & 0.3  & 0.6  & 0.4  & 0.5  & 58.6 & 63.9 & 61.0 \\
DS-Distilled-70B       & 0.0  & 0.0  & 0.0  & 0.0  & 0.0  & 0.0  & 0.0  & 0.0  & 0.0  & 57.8 & 62.3 & 59.9 \\
\rowcolor{blue!8}DS-Distilled-14B (SFT) & 65.3 & 72.7 & 67.1 & 37.5 & 40.5 & 38.1 & 65.3 & 72.7 & 67.1 & 95.7 & 94.3 & 94.9 \\
\rowcolor{blue!8}DS-Distilled-32B (SFT) & 74.6 & 77.1 & 75.5 & 44.5 & 46.2 & 45.1 & 74.6 & 77.1 & 75.5 & 98.3 & 97.1 & 97.6 \\
\rowcolor{blue!8}DS-Distilled-70B (SFT) & 57.4 & 43.4 & 46.1 & 34.3 & 26.5 & 27.8 & 57.4 & 43.4 & 46.1 & 80.0 & 81.4 & 80.1  \\
            \bottomrule
        \end{tabular}
    }
    \vspace{-0.3cm}
\end{table*}

\subsection{Legal Element Extraction}


\begin{table*}[!ht]
    \centering
    \caption{Performance of Legal Element Extraction tasks using 14B model.}
    \label{tab:legal_elements_14B}
    \renewcommand{\arraystretch}{1.2}
    \resizebox{\textwidth}{!}{
        \begin{tabular}{l|cccc|cccc|cccc}
            \toprule
            \multirow{2}{*}{\textbf{Model}} & \multicolumn{4}{c|}{\textbf{Beijing}} & \multicolumn{4}{c|}{\textbf{Tianjin}} & \multicolumn{4}{c}{\textbf{Hebei}} \\
            \cmidrule(lr){2-5} \cmidrule(lr){6-9} \cmidrule(lr){10-13}
            & Accuracy & Recall & Precision & F1 & Accuracy & Recall & Precision & F1 & Accuracy & Recall & Precision & F1 \\
            \midrule
            DS-Distilled-14B       & 33.5 & 36.3 & 84.9 & 49.1 & 32.4 & 35.5 & 80.5 & 48.0 & 31.8 & 35.3 & 80.0 & 47.5 \\
\rowcolor{blue!8}DS-Distilled-14B (SFT) & 85.0 & 89.5 & 94.3 & 91.5 & 84.9 & 87.0 & 97.0 & 91.3 & 82.7 & 87.2 & 93.9 & 89.9 \\
            \midrule
            & \multicolumn{4}{c|}{\textbf{Shanxi}} & \multicolumn{4}{c|}{\textbf{Nei Mongol}} & \multicolumn{4}{c}{\textbf{Liaoning}} \\
            \midrule
DS-Distilled-14B       & 16.4 & 20.0 & 47.5 & 26.9 & 23.9 & 27.5 & 63.1 & 36.6 & 16.7 & 20.4 & 47.4 & 27.2 \\
\rowcolor{blue!8}DS-Distilled-14B (SFT) & 54.2 & 63.4 & 72.8 & 66.6 & 78.1 & 82.9 & 92.0 & 86.6 & 40.3 & 52.1 & 60.7 & 55.0 \\
            \midrule
            & \multicolumn{4}{c|}{\textbf{Jilin}} & \multicolumn{4}{c|}{\textbf{Heilongjiang}} & \multicolumn{4}{c}{\textbf{Shanghai}} \\
            \midrule
DS-Distilled-14B       & 25.6 & 29.5 & 67.5 & 39.3 & 16.6 & 20.4 & 46.5 & 27.2 & 29.9 & 34.4 & 78.4 & 45.6 \\
\rowcolor{blue!8}DS-Distilled-14B (SFT) & 52.7 & 62.3 & 70.7 & 65.2 & 75.5 & 80.6 & 88.7 & 83.8 & 78.4 & 83.9 & 92.1 & 87.0 \\
            \midrule
            & \multicolumn{4}{c|}{\textbf{Jiangsu}} & \multicolumn{4}{c|}{\textbf{Zhejiang}} & \multicolumn{4}{c}{\textbf{Anhui}} \\
            \midrule
 DS-Distilled-14B       & 32.5 & 35.6 & 84.1 & 48.3 & 32.3 & 35.1 & 84.0 & 47.7 & 16.5 & 20.0 & 48.9 & 27.1 \\
\rowcolor{blue!8}DS-Distilled-14B (SFT) & 84.4 & 88.1 & 95.0 & 91.0 & 82.8 & 86.7 & 94.5 & 90.0 & 36.6 & 48.5 & 59.2 & 52.1 \\
            \midrule
            & \multicolumn{4}{c|}{\textbf{Fujian}} & \multicolumn{4}{c|}{\textbf{Jiangxi}} & \multicolumn{4}{c}{\textbf{Shandong}} \\
            \midrule
 DS-Distilled-14B       & 31.5 & 34.5 & 82.1 & 46.9 & 36.2 & 39.3 & 85.6 & 52.3 & 15.7 & 19.2 & 47.5 & 26.3 \\
\rowcolor{blue!8}DS-Distilled-14B (SFT) & 82.6 & 86.2 & 94.7 & 89.8 & 81.0 & 84.7 & 94.7 & 89.0 & 49.0 & 59.2 & 68.1 & 62.2 \\
            \midrule
            & \multicolumn{4}{c|}{\textbf{Henan}} & \multicolumn{4}{c|}{\textbf{Hubei}} & \multicolumn{4}{c}{\textbf{Hunan}} \\
            \midrule
DS-Distilled-14B       & 32.6 & 35.8 & 82.0 & 48.3 & 15.9 & 19.8 & 45.6 & 26.4 & 34.7 & 37.5 & 86.4 & 50.7 \\
\rowcolor{blue!8}DS-Distilled-14B (SFT) & 83.4 & 87.6 & 94.4 & 90.4 & 61.3 & 69.2 & 77.7 & 72.2 & 82.1 & 85.7 & 94.9 & 89.6 \\
            \midrule
            & \multicolumn{4}{c|}{\textbf{Guangdong}} & \multicolumn{4}{c|}{\textbf{Guangxi}} & \multicolumn{4}{c}{\textbf{Hainan}} \\
            \midrule
     DS-Distilled-14B       & 21.2 & 24.3 & 58.4 & 32.8 & 31.7 & 35.3 & 80.9 & 34.1 & 36.6 & 84.1 & 49.4 & 50.7 \\
\rowcolor{blue!8}DS-Distilled-14B (SFT) & 65.7 & 72.4 & 80.8 & 75.5 & 81.4 & 85.7 & 94.0 & 86.3 & 88.1 & 97.6 & 92.3 & 89.6 \\
            \midrule
            & \multicolumn{4}{c|}{\textbf{Chongqing}} & \multicolumn{4}{c|}{\textbf{Sichuan}} & \multicolumn{4}{c}{\textbf{Guizhou}} \\
            \midrule
  DS-Distilled-14B       & 29.4 & 32.3 & 82.6 & 45.0 & 31.4 & 34.1 & 82.2 & 46.7 & 31.8 & 35.5 & 80.0 & 47.4 \\
\rowcolor{blue!8}DS-Distilled-14B (SFT) & 77.9 & 82.8 & 92.4 & 86.8 & 80.3 & 84.9 & 93.6 & 88.4 & 82.1 & 86.2 & 94.5 & 89.7\\
            \midrule
            & \multicolumn{4}{c|}{\textbf{Yunnan}} & \multicolumn{4}{c|}{\textbf{Xizang}} & \multicolumn{4}{c}{\textbf{Shaanxi}} \\
            \midrule
DS-Distilled-14B       & 14.6 & 18.1 & 43.6 & 24.3 & 31.2 & 34.6 & 80.7 & 47.3 & 33.1 & 36.6 & 79.7 & 48.9 \\
\rowcolor{blue!8}DS-Distilled-14B (SFT) & 40.8 & 51.8 & 60.6 & 54.8 & 76.5 & 81.0 & 93.6 & 86.0 & 83.9 & 86.9 & 95.8 & 90.7 \\
            \midrule
            & \multicolumn{4}{c|}{\textbf{Gansu}} & \multicolumn{4}{c|}{\textbf{Qinghai}} & \multicolumn{4}{c}{\textbf{Ningxia}} \\
            \midrule
DS-Distilled-14B       & 26.7 & 30.6 & 67.3 & 40.4 & 35.2 & 37.8 & 85.2 & 51.2 & 35.5 & 37.5 & 90.0 & 51.4 \\
\rowcolor{blue!8}DS-Distilled-14B (SFT) & 52.4 & 61.6 & 69.8 & 64.4 & 84.3 & 87.8 & 95.4 & 91.0 & 83.2 & 85.4 & 96.8 & 90.4\\
            \midrule
            & \multicolumn{4}{c|}{\textbf{Xinjiang}} & \multicolumn{4}{c|}{\textbf{Average}} & \multicolumn{4}{c}{} \\
            \midrule
  DS-Distilled-14B       & 33.2 & 38.5 & 77.6 & 49.2 & 27.9 & 31.2 & 72.1 & 42.0 \\
\rowcolor{blue!8}DS-Distilled-14B (SFT) & 76.4 & 82.0 & 90.9 & 85.6 & 72.5 & 78.1 & 86.8 & 81.5 \\
            \bottomrule
        \end{tabular}
    }

\vspace{-1.5em}
\end{table*}

\begin{table*}[!ht]
    \centering
    \caption{Performance of Legal Element Extraction tasks using 32B model.}
    \label{tab:legal_elements_32B}
    \renewcommand{\arraystretch}{1.2}
    \resizebox{\textwidth}{!}{
        \begin{tabular}{l|cccc|cccc|cccc}
            \toprule
            \multirow{2}{*}{\textbf{Model}} & \multicolumn{4}{c|}{\textbf{Beijing}} & \multicolumn{4}{c|}{\textbf{Tianjin}} & \multicolumn{4}{c}{\textbf{Hebei}} \\
            \cmidrule(lr){2-5} \cmidrule(lr){6-9} \cmidrule(lr){10-13}
            & Accuracy & Recall & Precision & F1 & Accuracy & Recall & Precision & F1 & Accuracy & Recall & Precision & F1 \\
            \midrule
DS-Distilled-14B        & 35.8 & 38.1 & 89.5 & 51.0 & 36.8 & 40.7 & 81.5 & 53.1 & 37.3 & 40.8 & 83.8 & 53.4 \\
\rowcolor{blue!8}DS-Distilled-14B (SFT) & 83.0 & 88.8 & 92.3 & 90.1 & 82.2 & 85.5 & 95.4 & 89.7 & 81.5 & 86.3 & 93.3 & 89.2 \\
            \midrule
            & \multicolumn{4}{c|}{\textbf{Shanxi}} & \multicolumn{4}{c|}{\textbf{Nei Mongol}} & \multicolumn{4}{c}{\textbf{Liaoning}} \\
            \midrule
DS-Distilled-14B        & 24.7 & 28.3 & 61.6 & 37.3 & 20.7 & 24.3 & 55.8 & 32.3 & 16.8 & 20.8 & 46.2 & 27.2 \\
\rowcolor{blue!8}DS-Distilled-14B (SFT) & 44.3 & 54.8 & 65.3 & 58.3 & 68.6 & 75.6 & 84.5 & 79.0 & 41.6 & 53.4 & 61.7 & 56.2 \\
            \midrule
            & \multicolumn{4}{c|}{\textbf{Jilin}} & \multicolumn{4}{c|}{\textbf{Heilongjiang}} & \multicolumn{4}{c}{\textbf{Shanghai}} \\
            \midrule
DS-Distilled-14B        & 38.2 & 41.7 & 84.0 & 54.5 & 37.8 & 41.4 & 83.4 & 53.9 & 16.9 & 21.7 & 48.3 & 28.5 \\
\rowcolor{blue!8}DS-Distilled-14B (SFT) & 42.5 & 54.1 & 62.8 & 56.8 & 67.1 & 73.5 & 82.5 & 76.9 & 76.9 & 83.3 & 90.7 & 86.1\\
            \midrule
            & \multicolumn{4}{c|}{\textbf{Jiangsu}} & \multicolumn{4}{c|}{\textbf{Zhejiang}} & \multicolumn{4}{c}{\textbf{Anhui}} \\
            \midrule
 DS-Distilled-14B        & 24.5 & 28.6 & 63.1 & 37.7 & 37.7 & 40.7 & 87.6 & 53.7 & 17.1 & 20.7 & 49.4 & 28.0 \\
\rowcolor{blue!8}DS-Distilled-14B (SFT) & 83.3 & 87.6 & 94.1 & 90.4 & 81.8 & 86.7 & 93.2 & 89.3 & 38.8 & 50.3 & 60.6 & 53.8 \\
            \midrule
            & \multicolumn{4}{c|}{\textbf{Fujian}} & \multicolumn{4}{c|}{\textbf{Jiangxi}} & \multicolumn{4}{c}{\textbf{Shandong}} \\
            \midrule
DS-Distilled-14B        & 36.0 & 39.0 & 87.0 & 51.5 & 41.3 & 44.3 & 89.5 & 57.6 & 21.6 & 25.3 & 56.6 & 33.6 \\
\rowcolor{blue!8}DS-Distilled-14B (SFT) & 81.1 & 85.9 & 93.2 & 88.8 & 80.0 & 83.9 & 94.4 & 88.3 & 39.4 & 51.4 & 60.8 & 54.5\\
            \midrule
            & \multicolumn{4}{c|}{\textbf{Henan}} & \multicolumn{4}{c|}{\textbf{Hubei}} & \multicolumn{4}{c}{\textbf{Hunan}} \\
            \midrule
DS-Distilled-14B        & 33.8 & 37.3 & 77.4 & 48.9 & 37.7 & 41.4 & 84.5 & 53.7 & 25.7 & 30.1 & 61.4 & 38.8 \\
\rowcolor{blue!8}DS-Distilled-14B (SFT) & 81.8 & 86.8 & 93.3 & 89.4 & 71.1 & 77.4 & 86.0 & 80.8 & 80.7 & 84.5 & 94.5 & 88.8 \\
            \midrule
            & \multicolumn{4}{c|}{\textbf{Guangdong}} & \multicolumn{4}{c|}{\textbf{Guangxi}} & \multicolumn{4}{c}{\textbf{Hainan}} \\
            \midrule
DS-Distilled-14B        & 14.7 & 18.4 & 41.8 & 24.1 & 37.6 & 41.3 & 84.3 & 53.3 & 37.7 & 40.6 & 87.6 & 53.7 \\
\rowcolor{blue!8}DS-Distilled-14B (SFT) & 35.6 & 47.8 & 55.7 & 50.1 & 80.4 & 85.8 & 92.7 & 88.5 & 83.6 & 87.8 & 94.4 & 90.6\\
            \midrule
            & \multicolumn{4}{c|}{\textbf{Chongqing}} & \multicolumn{4}{c|}{\textbf{Sichuan}} & \multicolumn{4}{c}{\textbf{Guizhou}} \\
            \midrule
DS-Distilled-14B        & 31.8 & 33.8 & 89.1 & 46.5 & 36.7 & 40.1 & 85.6 & 52.9 & 35.4 & 39.9 & 75.7 & 50.8 \\
\rowcolor{blue!8}DS-Distilled-14B (SFT) & 78.4 & 83.3 & 92.3 & 87.1 & 79.1 & 84.5 & 92.3 & 87.7 & 81.0 & 85.7 & 93.5 & 88.9\\
            \midrule
            & \multicolumn{4}{c|}{\textbf{Yunnan}} & \multicolumn{4}{c|}{\textbf{Xizang}} & \multicolumn{4}{c}{\textbf{Shaanxi}} \\
            \midrule
DS-Distilled-14B        & 15.9 & 20.0 & 44.4 & 26.3 & 40.3 & 42.8 & 89.2 & 55.7 & 20.1 & 24.5 & 51.9 & 31.6 \\
\rowcolor{blue!8}DS-Distilled-14B (SFT) & 32.8 & 45.2 & 54.0 & 48.0 & 80.7 & 83.4 & 95.9 & 88.8 & 81.7 & 85.7 & 94.3 & 89.4 \\
            \midrule
            & \multicolumn{4}{c|}{\textbf{Gansu}} & \multicolumn{4}{c|}{\textbf{Qinghai}} & \multicolumn{4}{c}{\textbf{Ningxia}} \\
            \midrule
DS-Distilled-14B        & 18.3 & 22.6 & 48.0 & 29.3 & 28.4 & 32.8 & 67.9 & 42.5 & 39.1 & 41.3 & 91.7 & 55.6 \\
\rowcolor{blue!8}DS-Distilled-14B (SFT) & 42.3 & 53.2 & 63.0 & 56.6 & 83.5 & 87.4 & 94.9 & 90.5 & 80.9 & 84.1 & 95.2 & 88.9\\
            \midrule
            & \multicolumn{4}{c|}{\textbf{Xinjiang}} & \multicolumn{4}{c|}{\textbf{Average}} & \multicolumn{4}{c}{} \\
            \midrule
DS-Distilled-14B        & 41.6 & 47.1 & 81.0 & 57.7 & 30.3 & 33.9 & 71.9 & 44.3 \\
\rowcolor{blue!8}DS-Distilled-14B (SFT) & 77.9 & 83.5 & 91.6 & 86.8 & 69.2 & 75.7 & 84.2 & 79.0\\
            \bottomrule
        \end{tabular}
    }
\end{table*}



We construct a high-quality dataset specifically focusing on the ``crime of intentional injury'' for the legal element extraction task. This dataset comprises over $50,000$ training samples and more than $10,000$ test samples, following an $8:2$ train-test split. To evaluate the effectiveness of supervised fine-tuning (SFT), we apply our dataset to fine-tune models of two scales: 14B and 32B. As shown in Table~\ref{tab:legal_elements_14B} and Table~\ref{tab:legal_elements_32B}, we present detailed evaluation metrics including accuracy, recall, precision, and F1 scores for each of the 31 provinces, alongside their average values. The results demonstrate significant performance gains after supervised fine-tuning, particularly highlighting improvements exceeding 40\% in critical metrics such as accuracy, recall, and F1 scores. These substantial improvements confirm both the necessity and the effectiveness of fine-tuning for domain-specific legal tasks. Moreover, the impressive performance gains underline the importance of utilizing high-quality, carefully annotated datasets, as they directly contribute to the model’s ability to accurately recognize and extract complex legal elements.

\begin{figure*}[!t]
\centering
\includegraphics[width=0.9\linewidth]{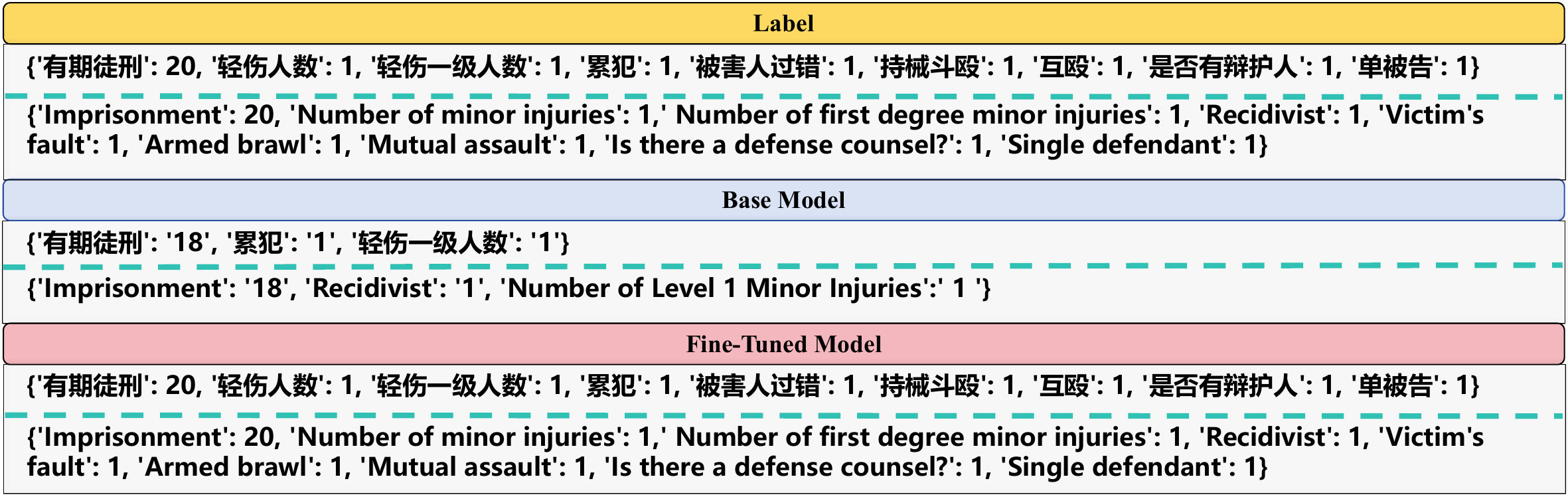}
\caption{Example of legal extraction task inference results. The base model fails to capture several legal elements, while the fine-tuned model successfully aligns with the ground truth labels.}
\label{fig:legal_element_extraction_1}
\end{figure*}

\begin{figure*}[!t]
\centering
\includegraphics[width=0.9\linewidth]{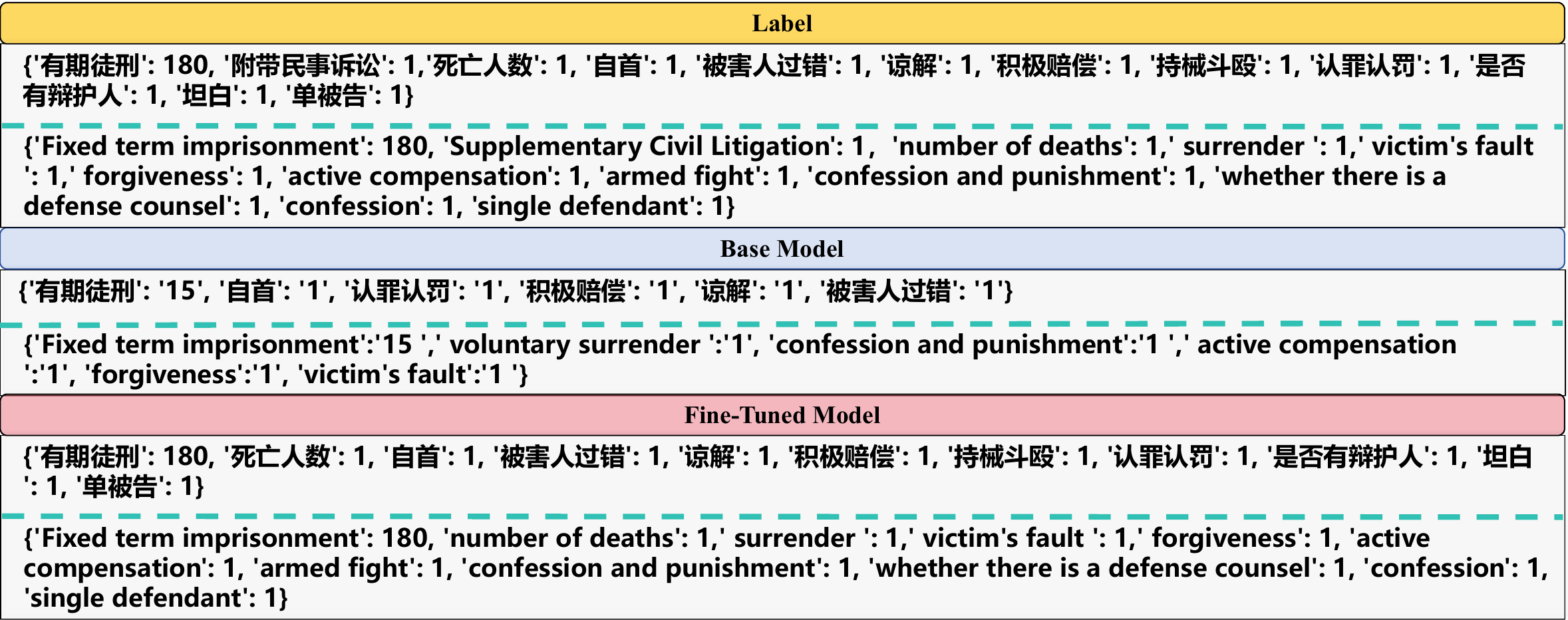}
\caption{Example of legal extraction task inference results. The base model misses several legal elements and uses inconsistent units (e.g., "15 years" instead of the expected "180 months"). While the fine-tuned model still omits one element (civil litigation), it standardizes the unit representation to months.}
\label{fig:legal_element_extraction_2}
\vspace{-0.5em}
\end{figure*}

\begin{figure*}[!b]
\centering
\includegraphics[width=0.9\linewidth]{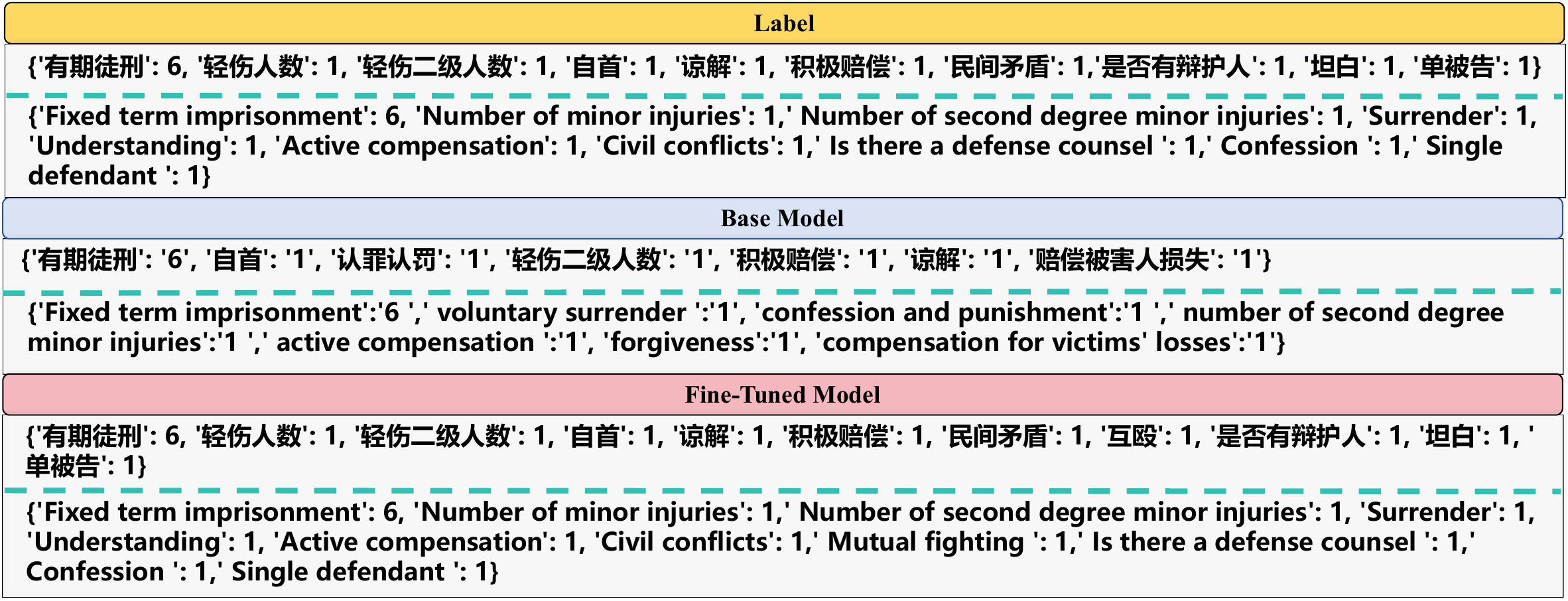}
\caption{Example of legal extraction task inference results. The base model fails to output the total number of minor injuries when predicting the number of Level 2 minor injuries, resulting in missing elements. The fine-tuned model correctly identifies all required elements but introduces an additional "mutual fighting" element.}
\label{fig:legal_element_extraction_3}
\vspace{-0.5em}
\end{figure*}

\begin{figure*}[!t]
\centering
\includegraphics[width=0.9\linewidth]{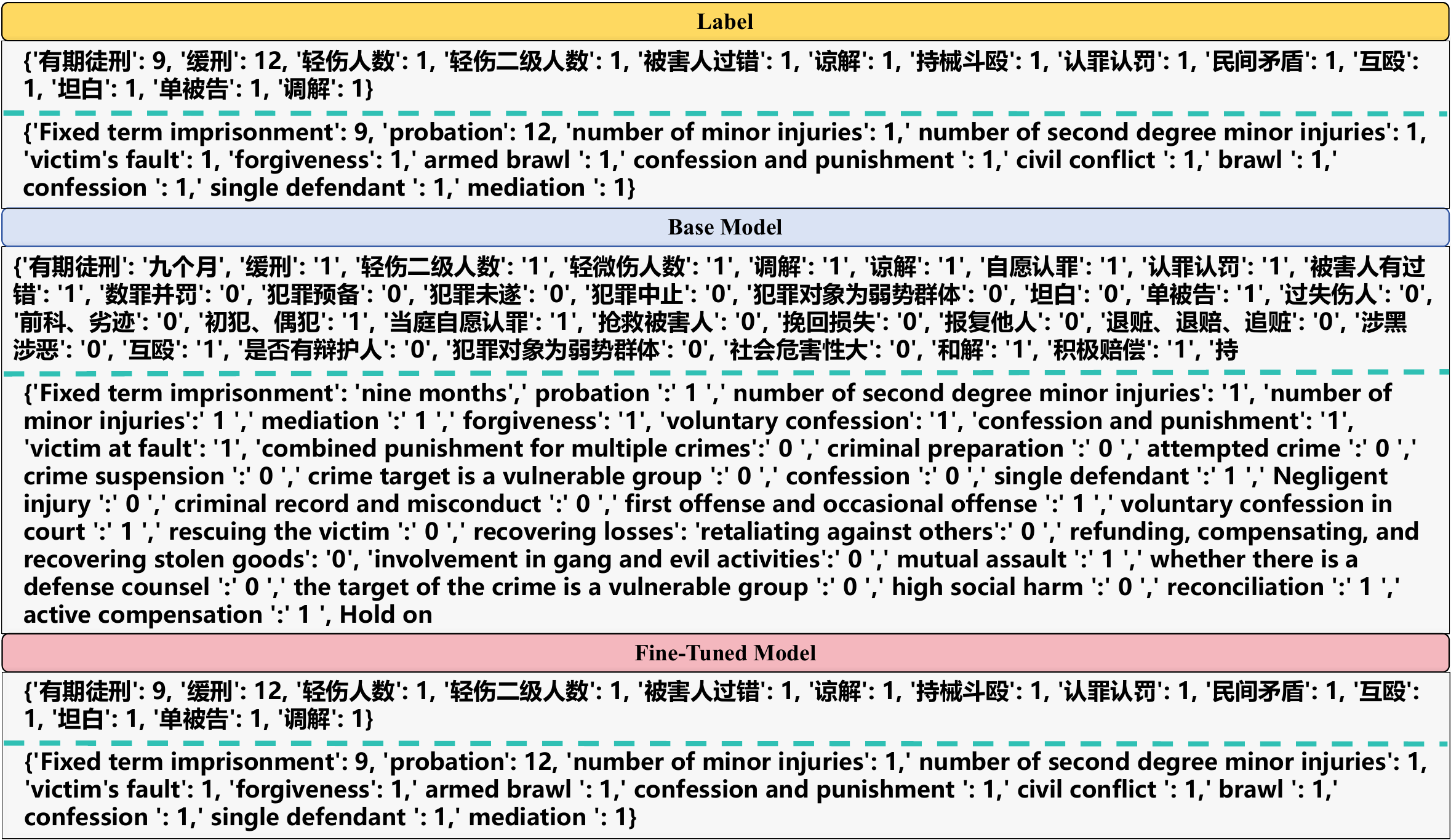}
\caption{Example of legal extraction task inference results. The base model does not follow the required output format, while the fine-tuned model produces fully correct results.}
\label{fig:legal_element_extraction_4}
\end{figure*}

\begin{figure*}[!t]
\centering
\includegraphics[width=0.9\linewidth]{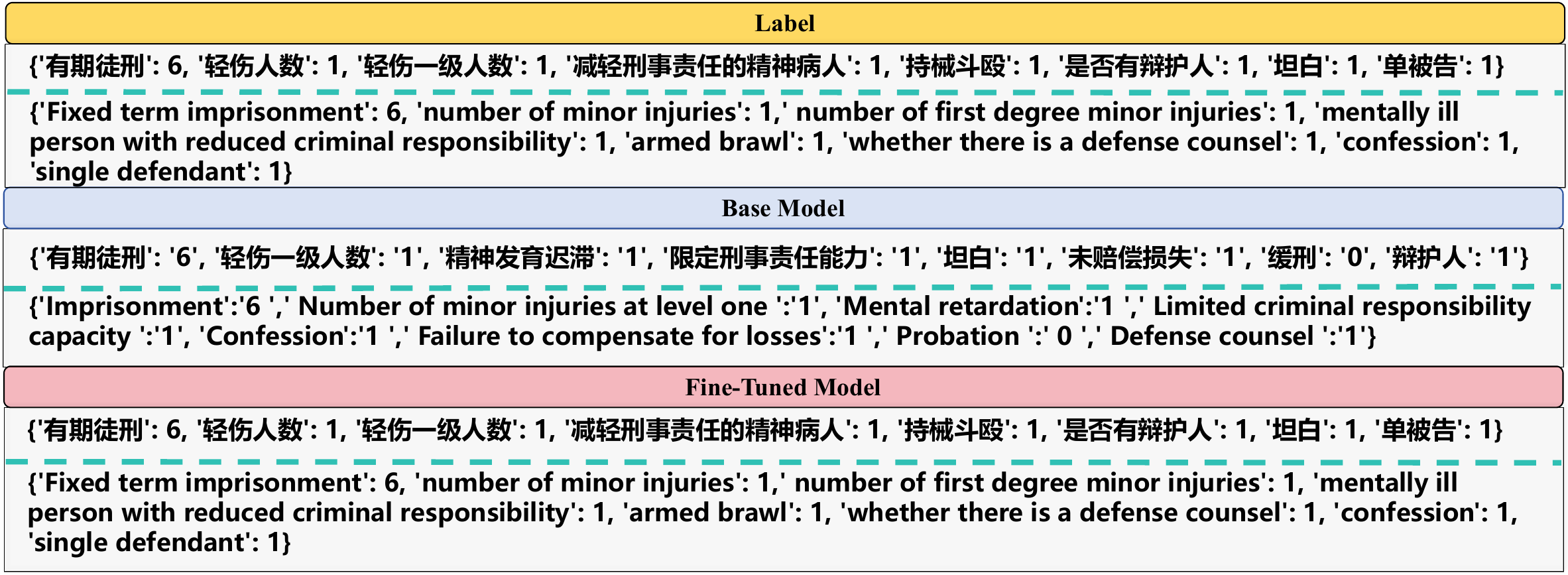}
\caption{Example of legal extraction task inference results. The base model fails to follow the required element constraints and outputs unrecognized elements. After fine-tuning, all elements are correctly identified.}
\label{fig:legal_element_extraction_5}
\end{figure*}

\begin{figure*}[!t]
\centering
\includegraphics[width=0.9\linewidth]{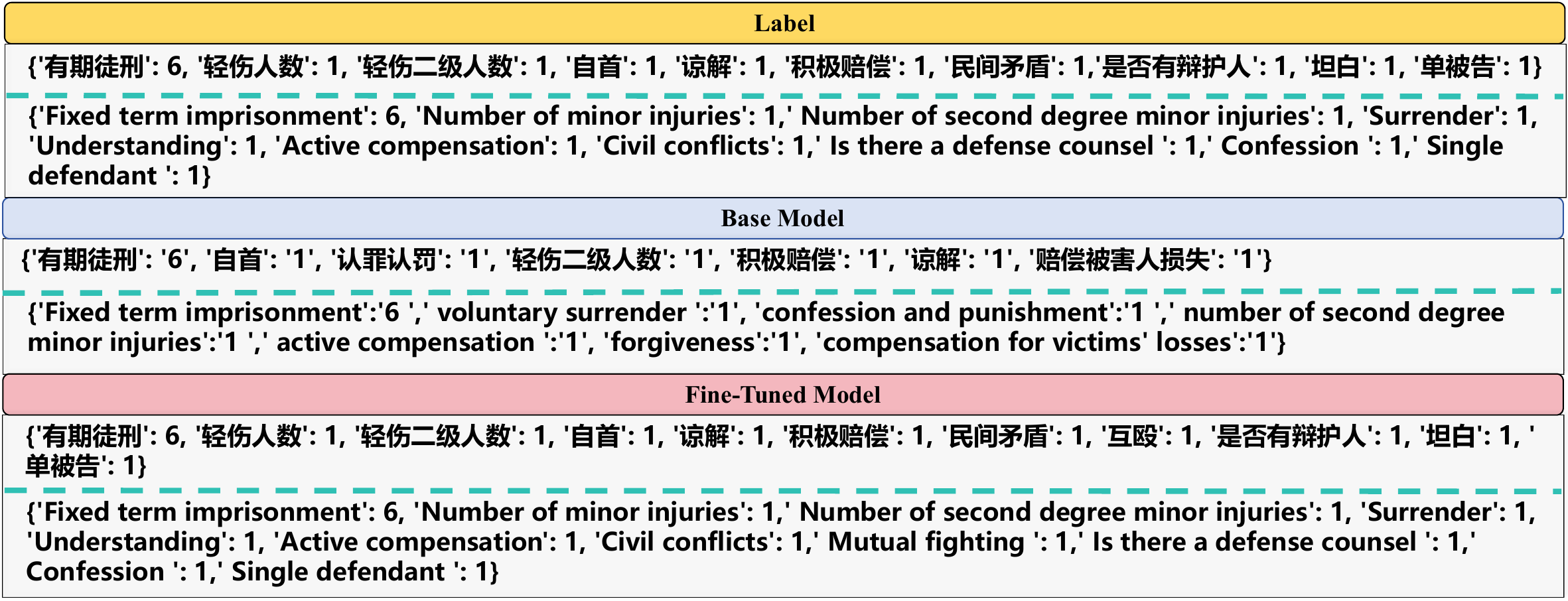}
\caption{Example of legal extraction task inference results. The base model produces inconsistent and incorrect outputs for imprisonment duration, while the fine-tuned model correctly extracts the date.}
\label{fig:legal_element_extraction_6}
\end{figure*}

\begin{figure*}[!t]
\centering
\includegraphics[width=0.9\linewidth]{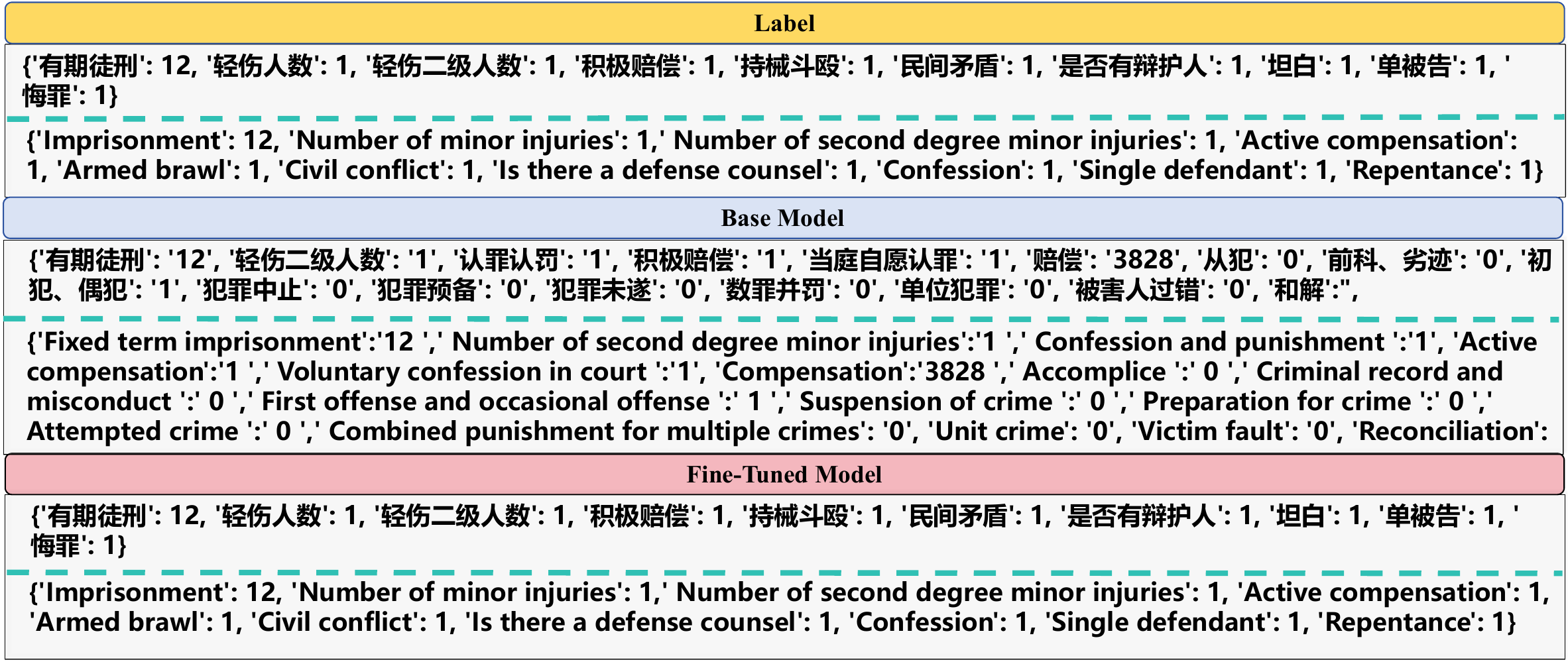}
\caption{Example of legal extraction task inference results. The base model fails to output the number of minor injuries, generates nonexistent elements, and does not follow instructions. After fine-tuning, all outputs are correct.}
\label{fig:legal_element_extraction_7}
\end{figure*}

\begin{figure*}[!t]
\centering
\includegraphics[width=0.9\linewidth]{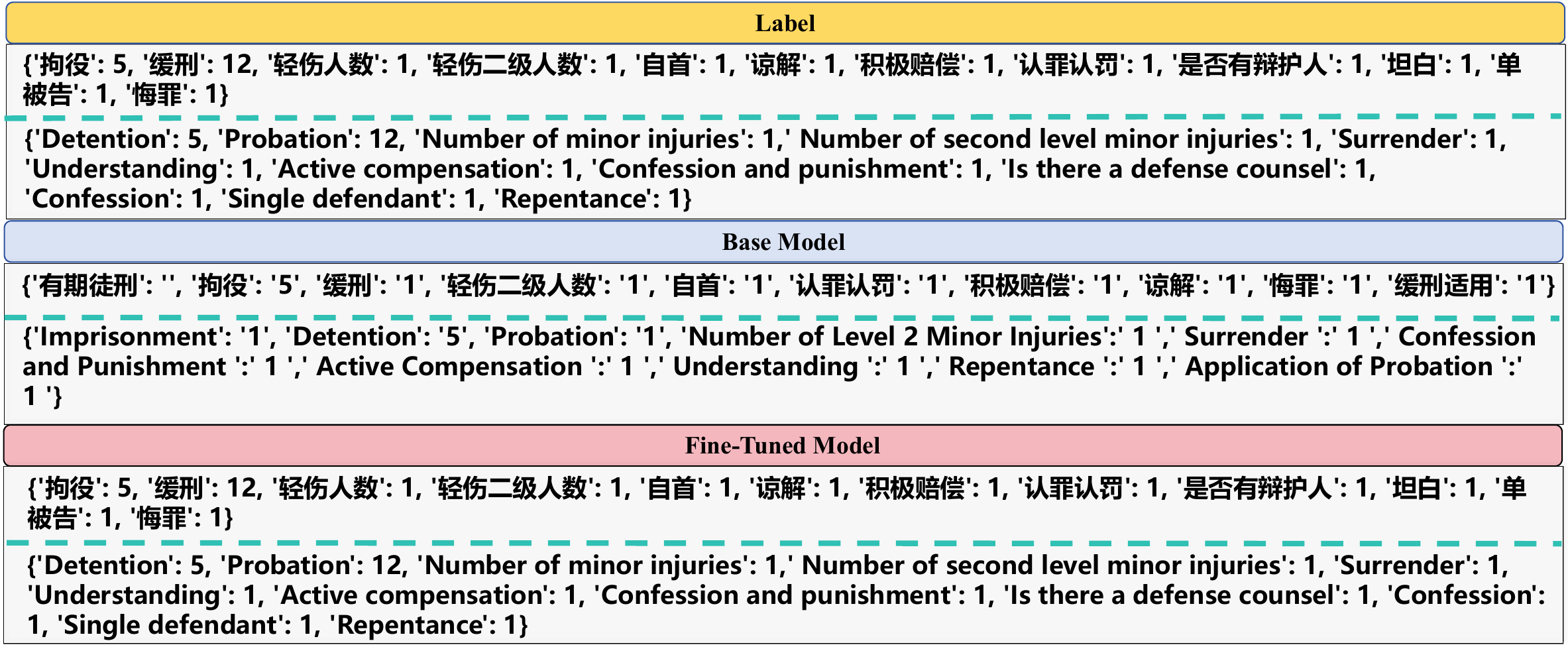}
\caption{Example of legal extraction task inference results. The base model outputs unrecognized elements like imprisonment, uses incorrect units for probation, fabricates elements, and fails to fully recognize required elements. After fine-tuning, all outputs are correct.}
\label{fig:legal_element_extraction_8}
\end{figure*}

\begin{figure*}[!t]
\centering
\includegraphics[width=0.9\linewidth]{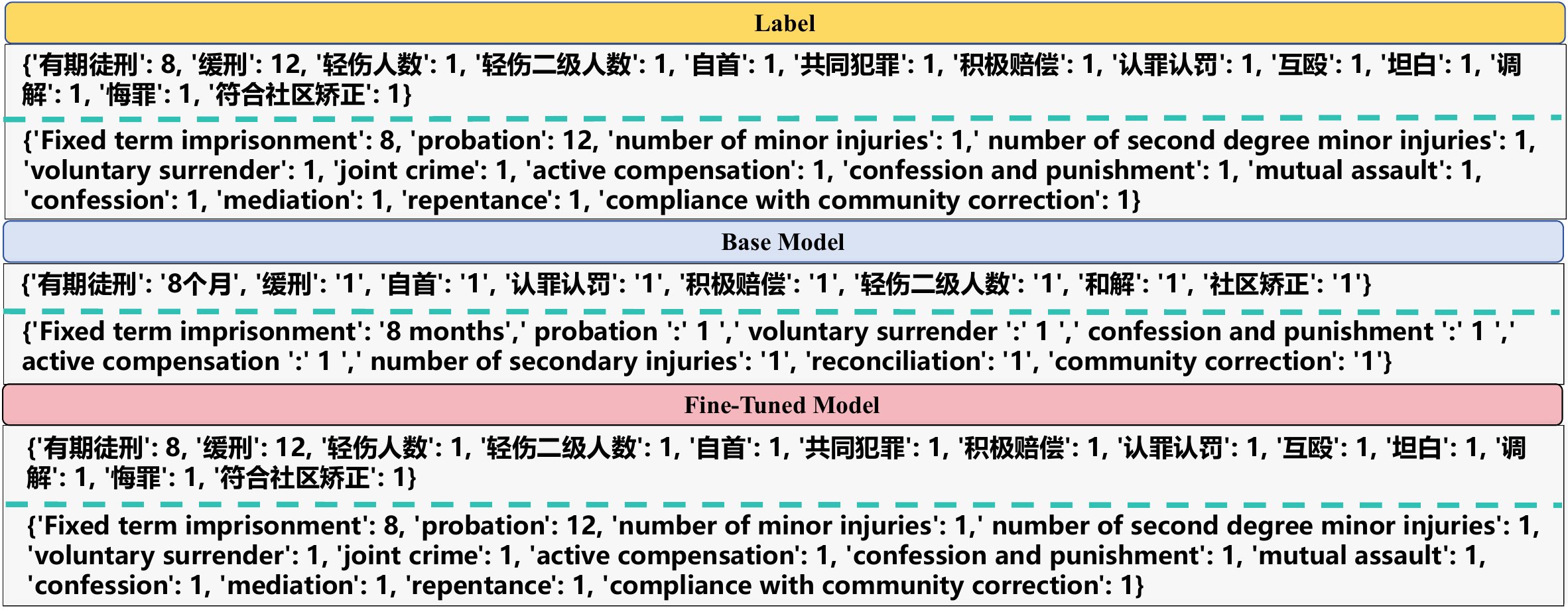}
\caption{Example of legal extraction task inference results. The base model fails to follow the specified format, uses inconsistent units, alters element names (e.g., changing "compliance with community correction" to "community correction"), and fails to identify all required elements. After fine-tuning, all outputs are correct.}
\label{fig:legal_element_extraction_9}
\end{figure*}

\begin{figure*}[!t]
\centering
\includegraphics[width=0.9\linewidth]{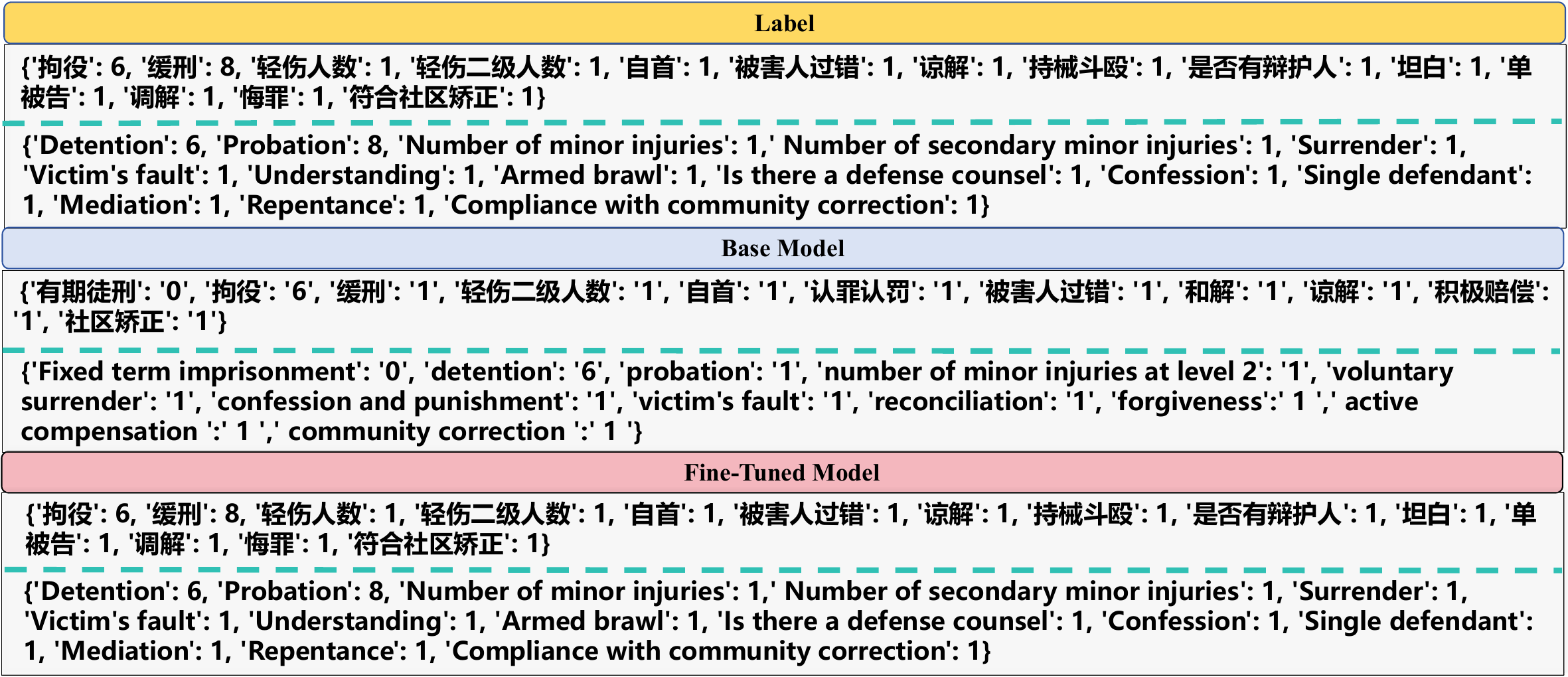}
\caption{Example of legal extraction task inference results. The base model outputs nonexistent elements, misses required elements, and modifies element names (e.g., changing "compliance with community correction" to "community correction"). After fine-tuning, all outputs are correct.}
\label{fig:legal_element_extraction_10}
\end{figure*}

\begin{figure*}[!t]
\centering
\includegraphics[width=0.9\linewidth]{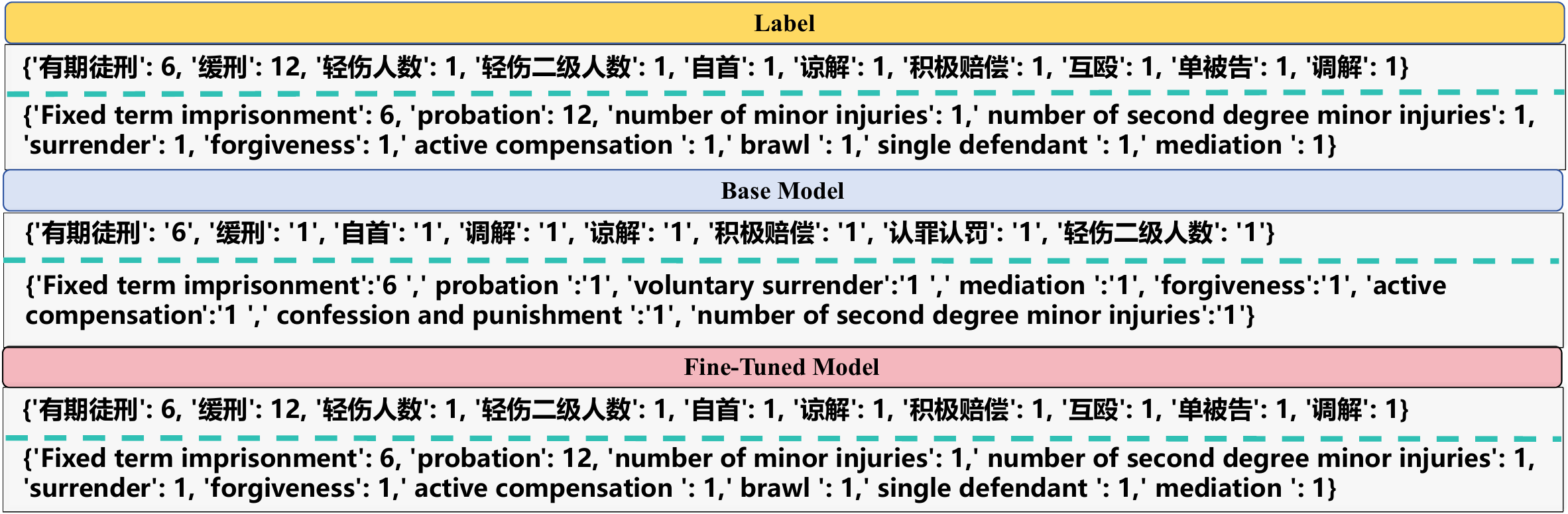}
\caption{Example of legal extraction task inference results. The base model incorrectly outputs probation details, only provides the number of Level 2 minor injuries without categorizing them under total minor injuries, and fails to recognize some elements. After fine-tuning, all outputs are correct.}
\label{fig:legal_element_extraction_11}
\end{figure*}

\begin{figure*}[!t]
\centering
\includegraphics[width=0.9\linewidth]{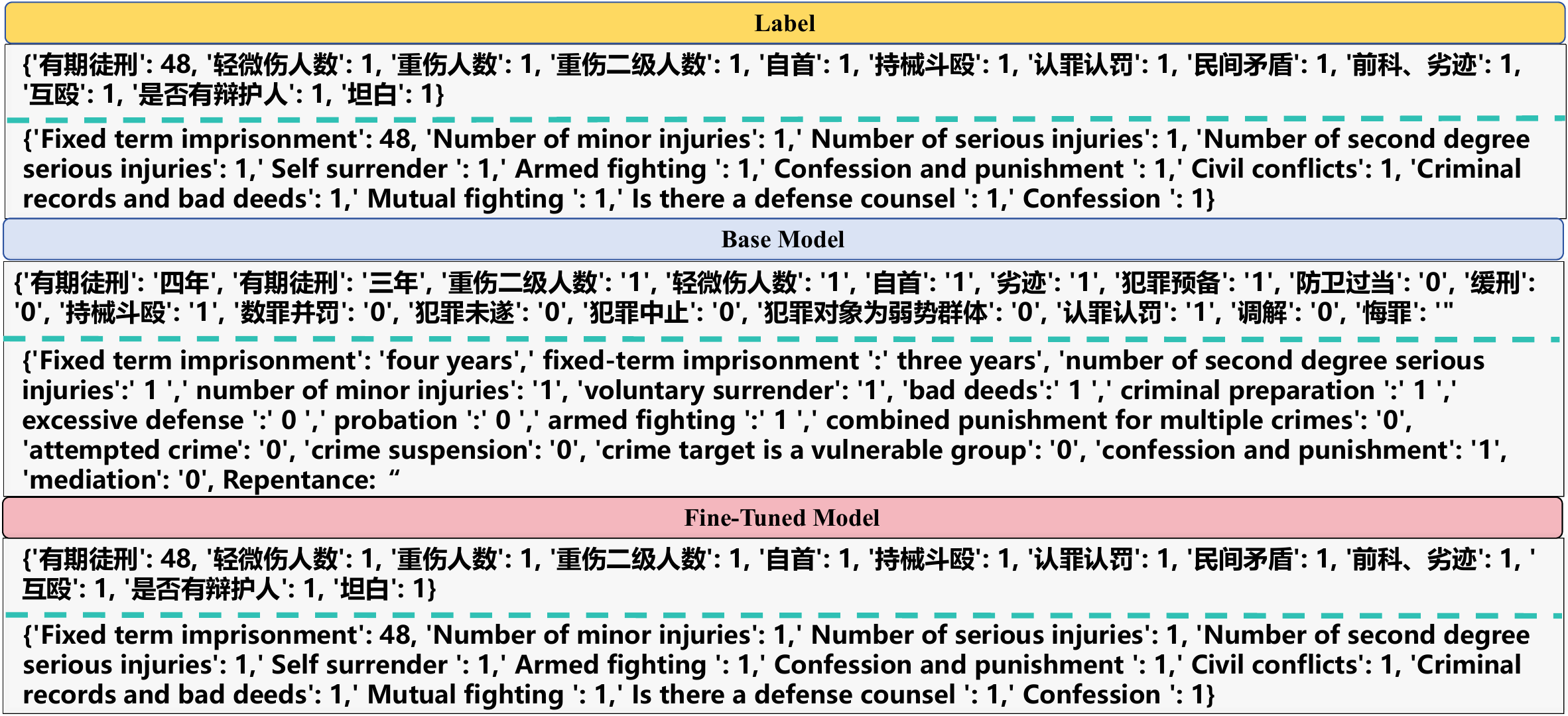}
\caption{Example of legal extraction task inference results. The base model incorrectly outputs imprisonment details, fails to recognize the number of serious injuries, and generates nonexistent elements. After fine-tuning, all outputs are correct.}
\label{fig:legal_element_extraction_12}
\end{figure*}

\begin{figure*}[!t]
\centering
\includegraphics[width=0.9\linewidth]{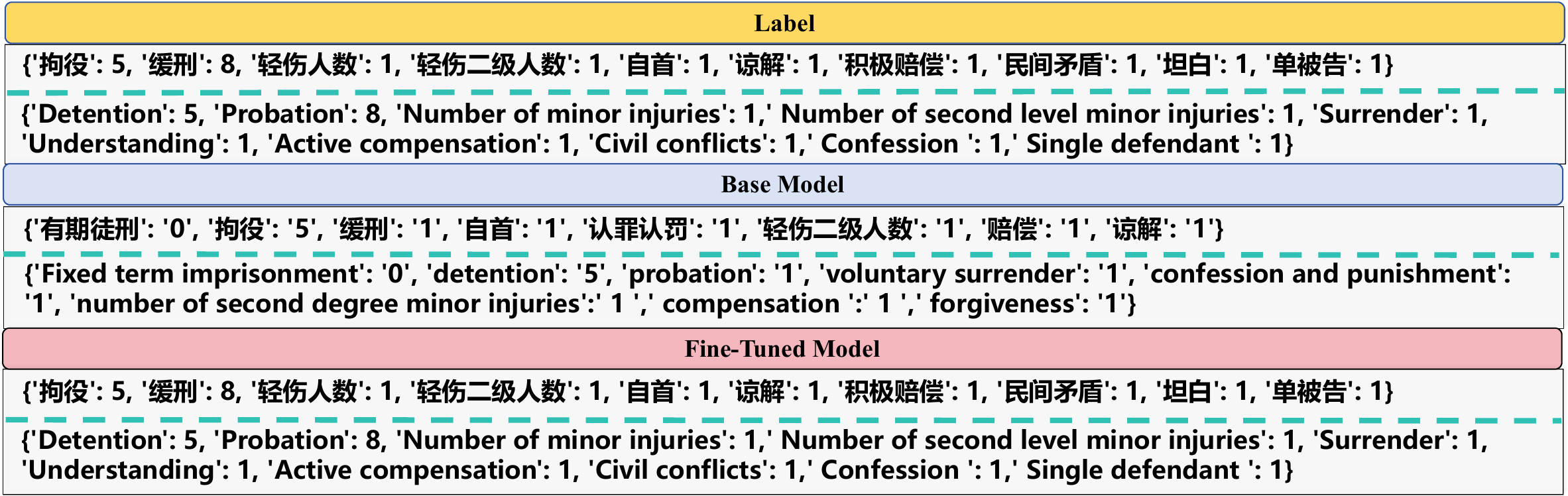}
\caption{Example of legal extraction task inference results. The base model incorrectly outputs the imprisonment element despite its absence, misses required elements, and modifies element names (e.g., changing "compensation"). After fine-tuning, all outputs are correct.}
\label{fig:legal_element_extraction_13}
\end{figure*}

\begin{figure*}[!t]
\centering
\includegraphics[width=0.9\linewidth]{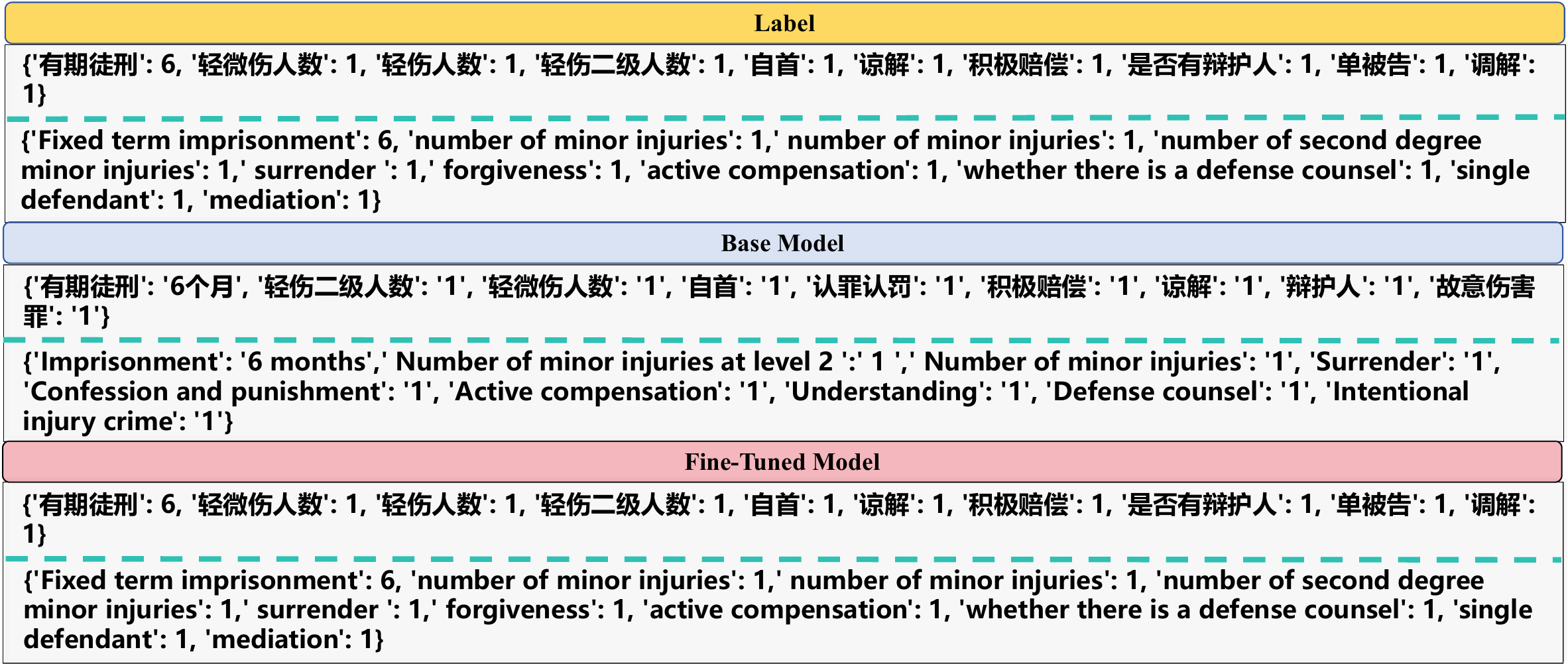}
\caption{Example of legal extraction task inference results. The base model generates element names that are not in the predefined list and outputs nonexistent elements. After fine-tuning, all outputs are correct.}
\label{fig:legal_element_extraction_14}
\end{figure*}

\begin{figure*}[!t]
\centering
\includegraphics[width=0.9\linewidth]{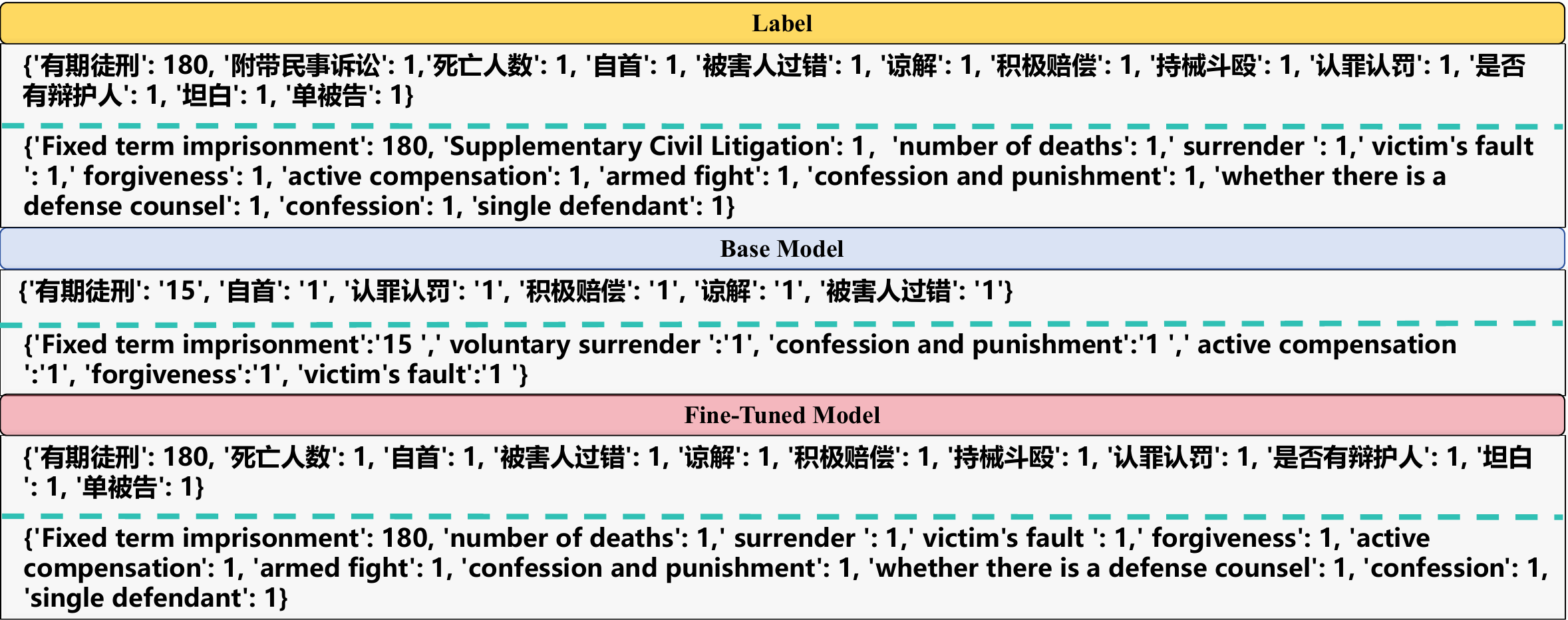}
\caption{Example of legal extraction task inference results. The base model does not follow the required format in its output. After fine-tuning, both the format and results are entirely correct.}
\label{fig:legal_element_extraction_15}
\end{figure*}

\begin{figure*}[!t]
\centering
\includegraphics[width=0.9\linewidth]{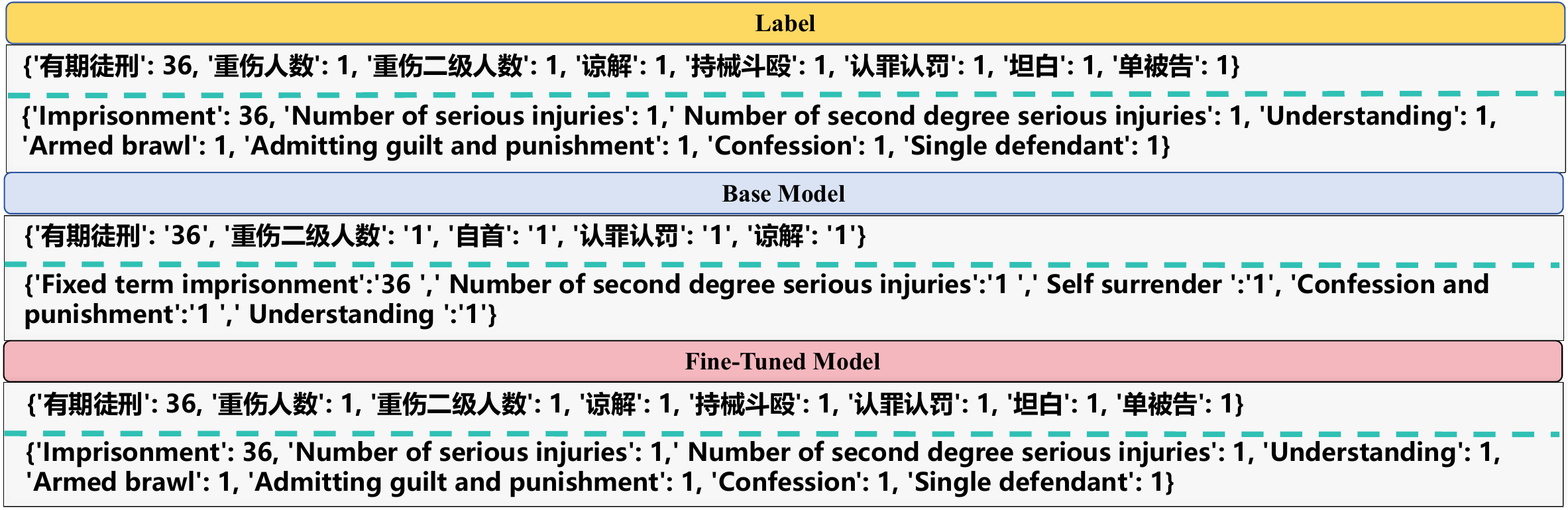}
\caption{Example of legal extraction task inference results. The base model fails to recognize the number of serious injuries, incorrectly outputs the element "self surrender",  and does not fully extract all required elements. After fine-tuning, all outputs are correct.}
\label{fig:legal_element_extraction_16}
\end{figure*}

\begin{figure*}[!t]
\centering
\includegraphics[width=0.9\linewidth]{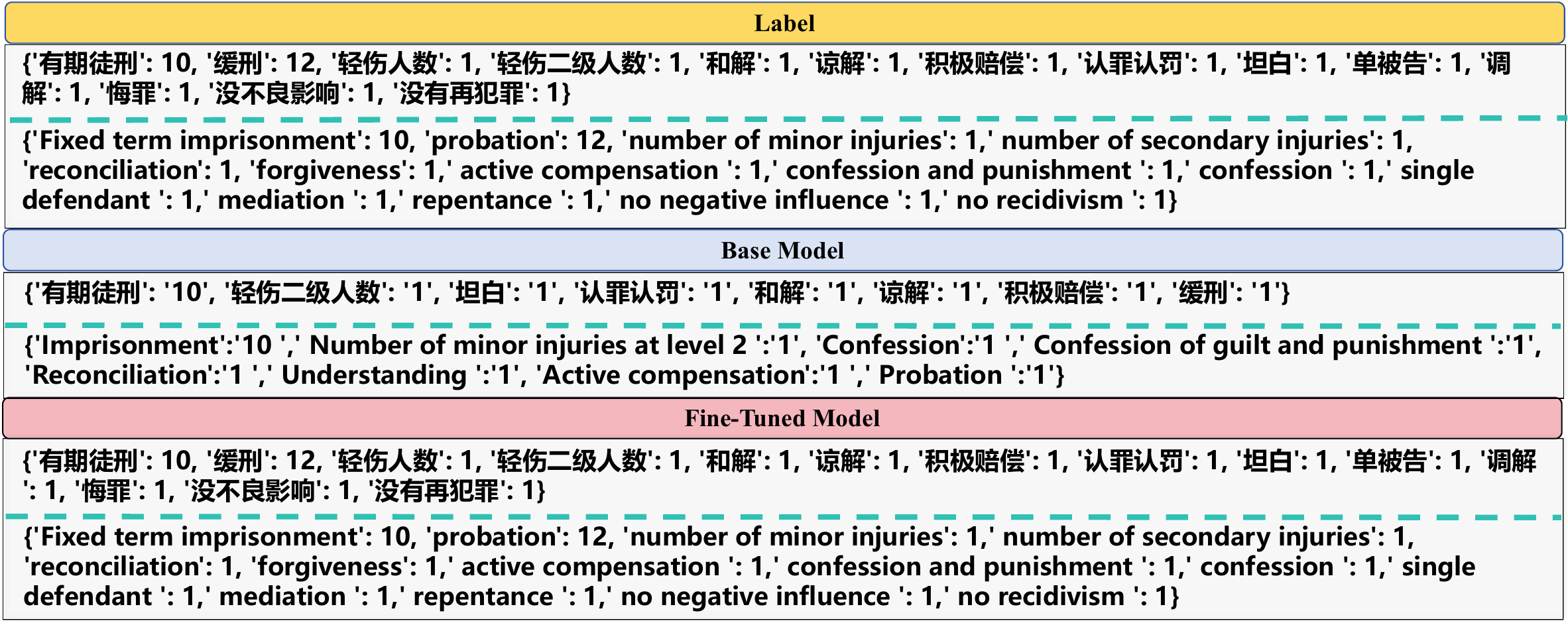}
\caption{Example of legal extraction task inference results. The base model misidentifies the unit for probation, fails to recognize the number of minor injuries, and misses many required elements. After fine-tuning, all outputs are correct.}
\label{fig:legal_element_extraction_17}
 \vspace{-2.5em}
\end{figure*}

\begin{figure*}[!t]
\centering
\includegraphics[width=0.9\linewidth]{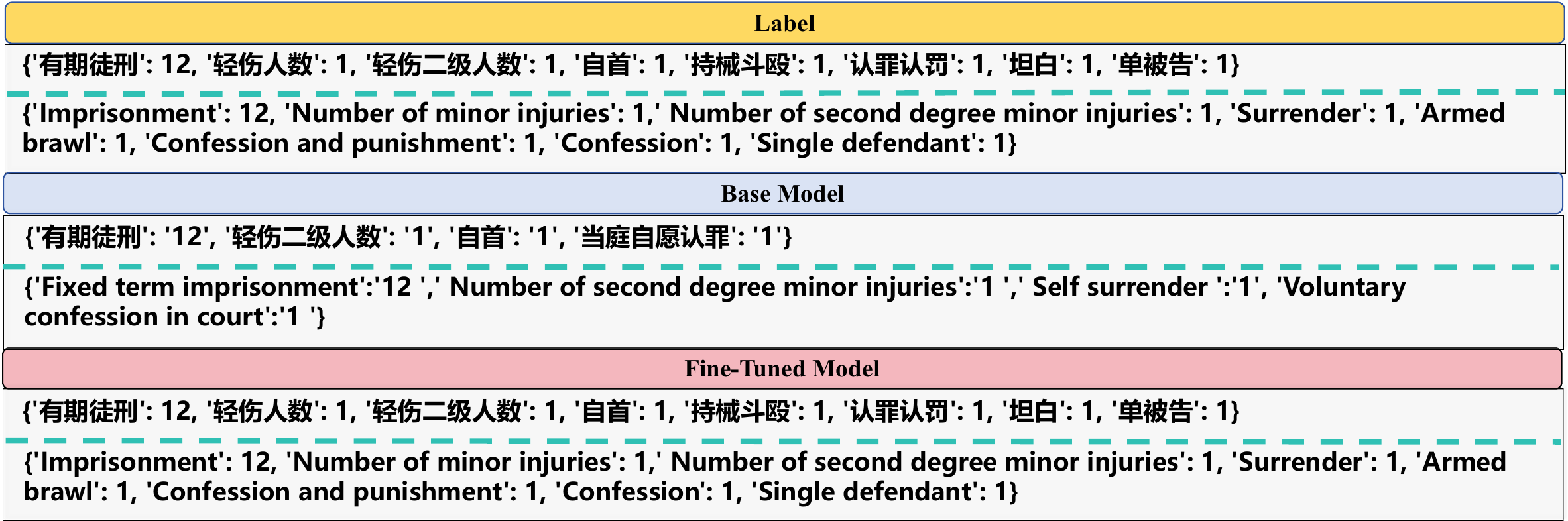}
\caption{Example of legal extraction task inference results. The base model misidentifies the unit for probation, fails to recognize the number of minor injuries, and misses many required elements. After fine-tuning, all outputs are correct.}
\label{fig:legal_element_extraction_18}
 \vspace{-2.5em}
\end{figure*}

\begin{figure*}[!t]
\centering
\includegraphics[width=0.9\linewidth]{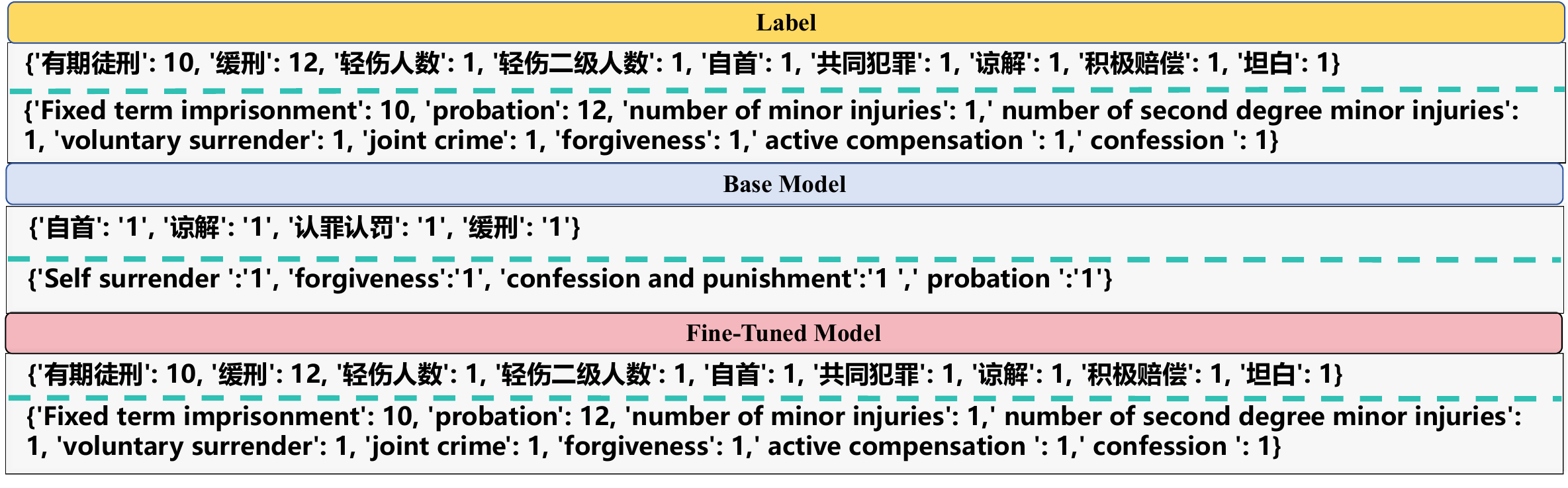}
\caption{Example of legal extraction task inference results. The base model misses elements such as imprisonment and incorrectly outputs the "confession and punishment" element.}
\label{fig:legal_element_extraction_19}
 \vspace{-2.5em}
\end{figure*}

\begin{figure*}[!t]
\centering
\includegraphics[width=0.9\linewidth]{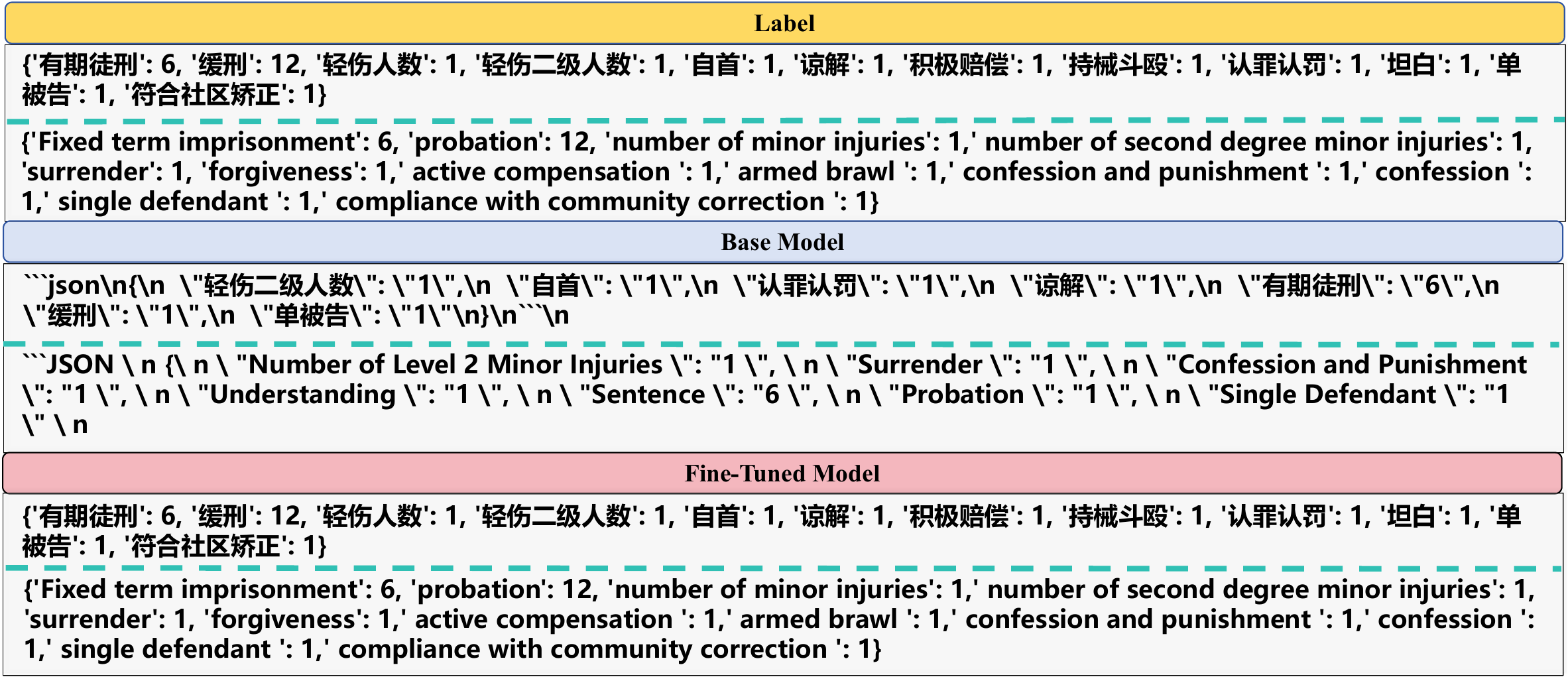}
\caption{Example of legal extraction task inference results. The base model produces highly disorganized output, resulting in unreadable text.}
\label{fig:legal_element_extraction_20}
\end{figure*}

\begin{figure*}[!t]
\centering
\includegraphics[width=0.9\linewidth]{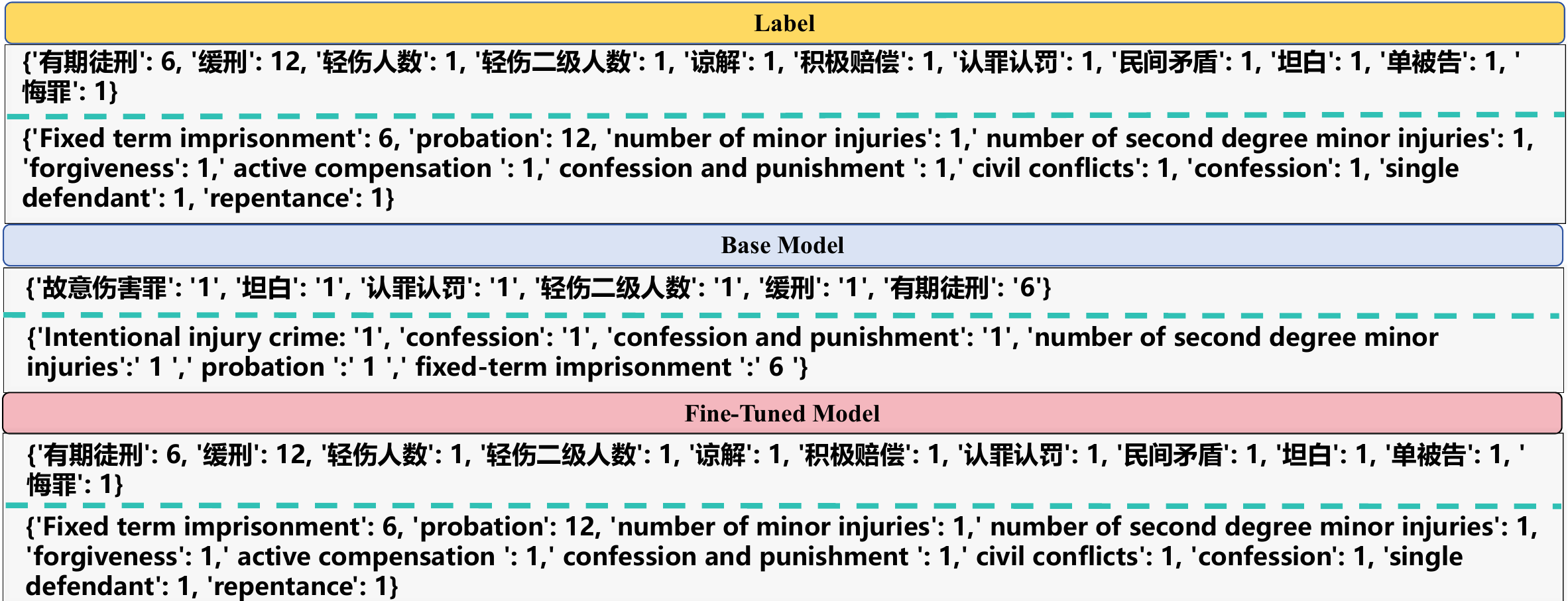}
\caption{Example of legal extraction task inference results. The base model uses an incorrect unit for probation, fails to recognize the number of minor injuries, and misses additional required elements. After fine-tuning, all outputs are correct.}
\label{fig:legal_element_extraction_21}
\end{figure*}

\begin{figure*}[!t]
\centering
\includegraphics[width=0.9\linewidth]{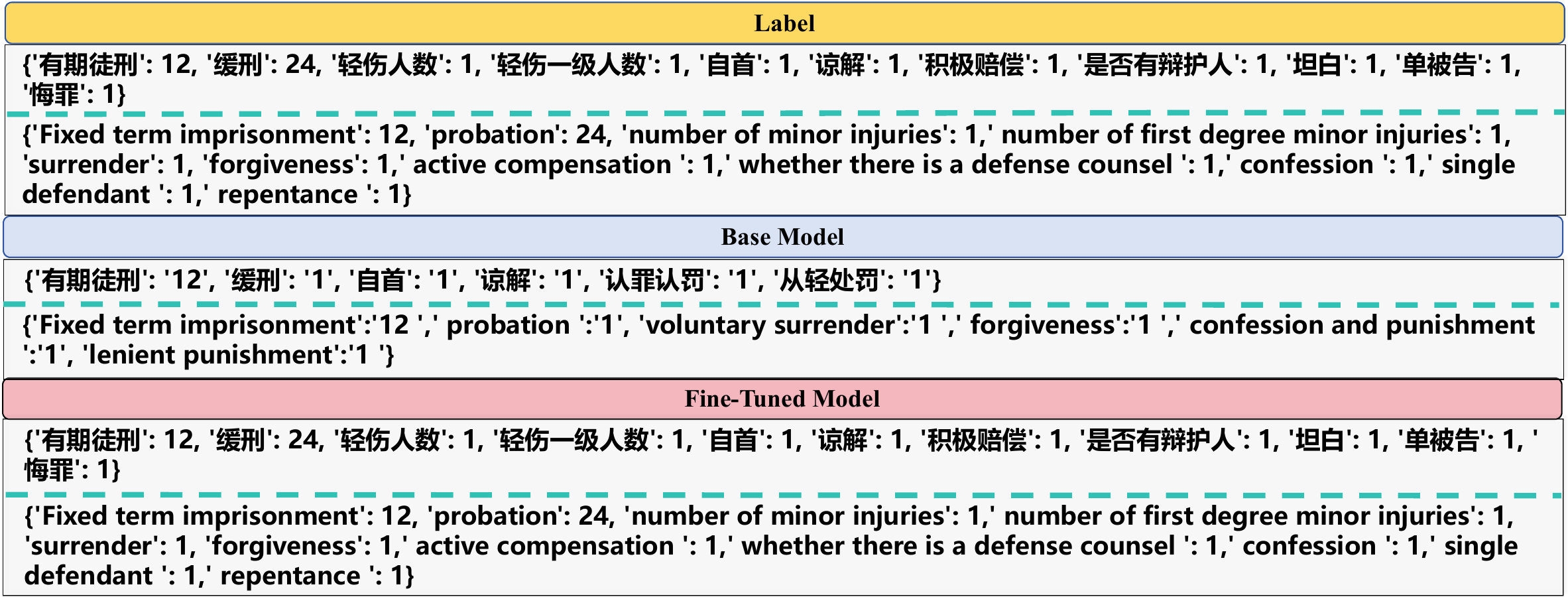}
\caption{Example of legal extraction task inference results. The base model incorrectly identifies probation and fails to extract many required elements. After fine-tuning, all outputs are correct.}
\label{fig:legal_element_extraction_22}
\end{figure*}

\begin{figure*}[!t]
\centering
\includegraphics[width=0.9\linewidth]{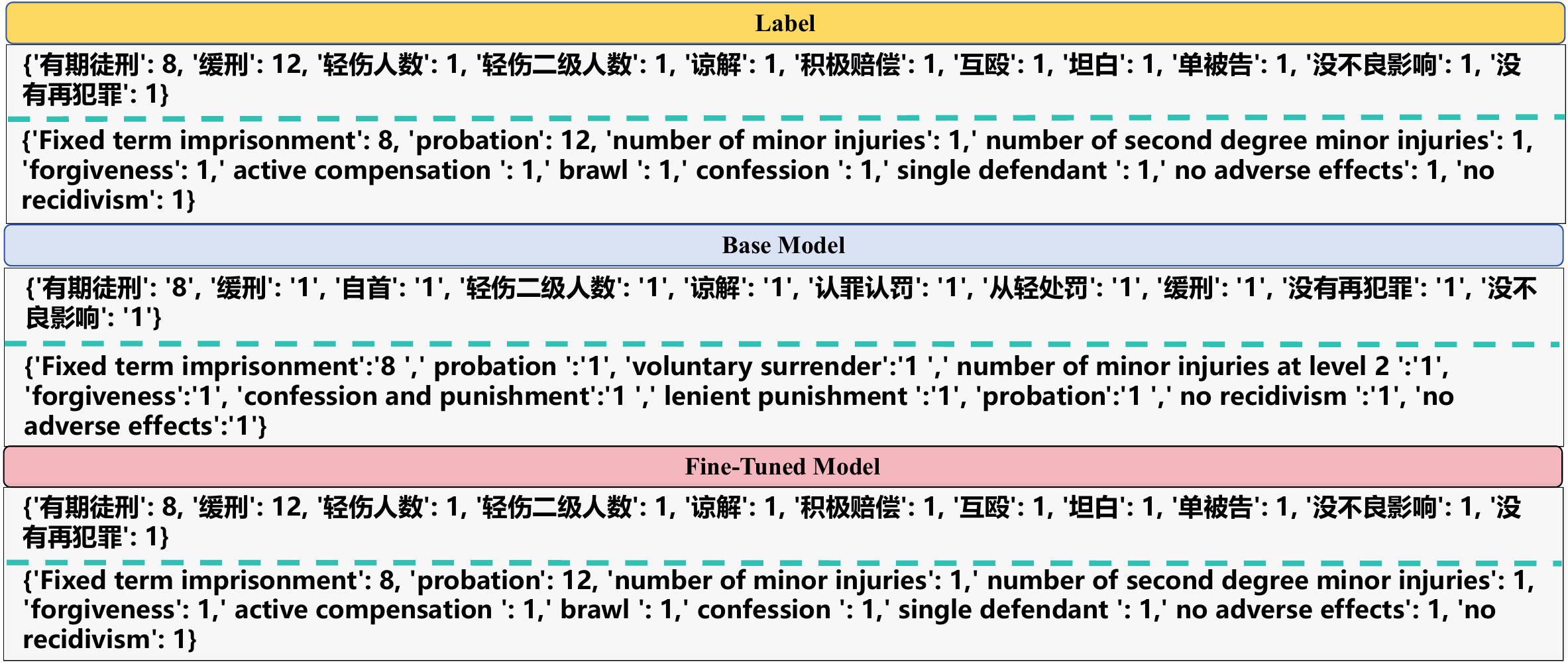}
\caption{Example of legal extraction task inference results. The base model misses required elements, outputs nonexistent elements (e.g., "no adverse effects"), fails to recognize the number of minor injuries, and uses an incorrect unit for probation. After fine-tuning, all outputs are correct.}
\label{fig:legal_element_extraction_23}
\end{figure*}

\begin{figure*}[!t]
\centering
\includegraphics[width=0.9\linewidth]{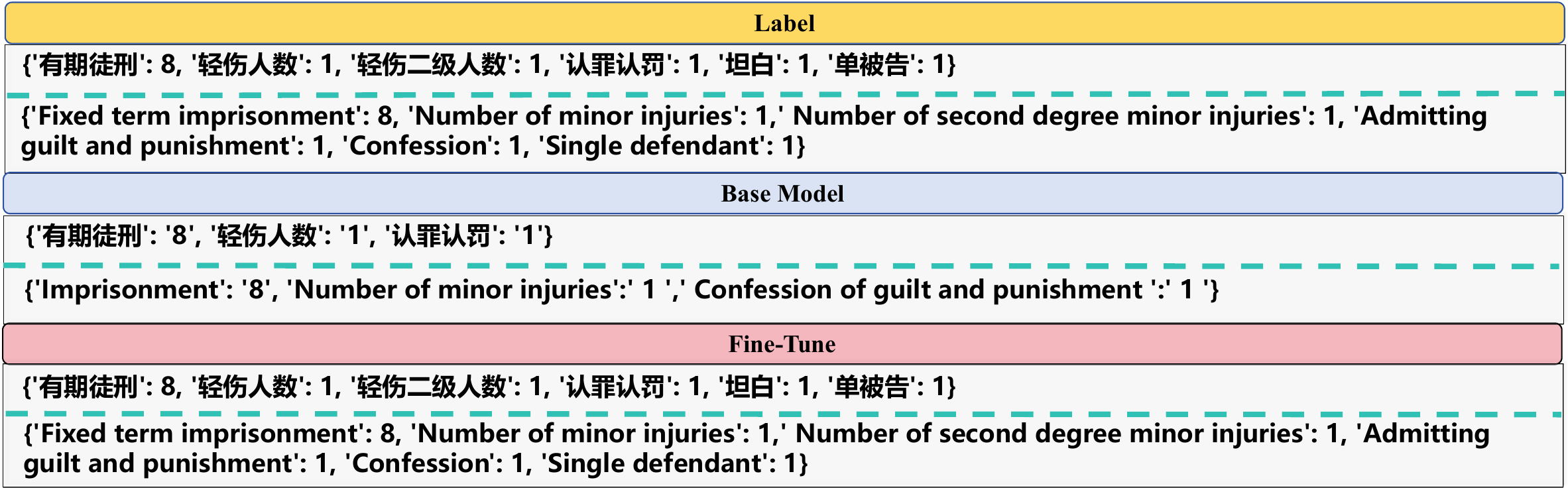}
\caption{Example of legal extraction task inference results. The base model fails to recognize some required elements. After fine-tuning, all outputs are correct.}
\label{fig:legal_element_extraction_24}
\end{figure*}

\begin{figure*}[!t]
\centering
\includegraphics[width=0.9\linewidth]{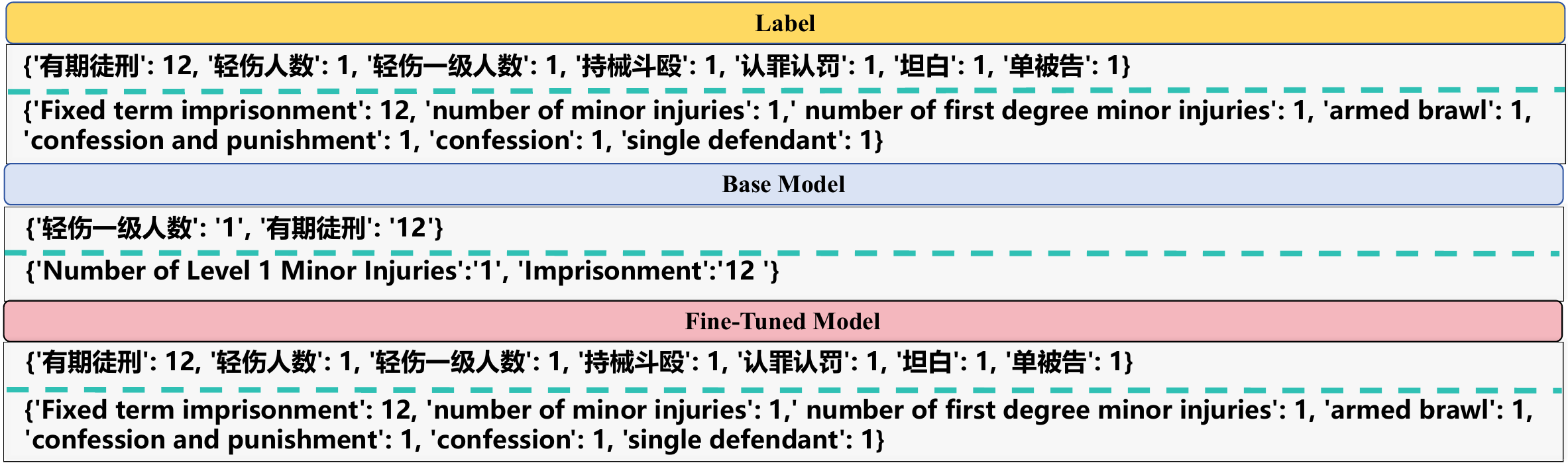}
\caption{Example of legal extraction task inference results. The base model fails to output many required elements. After fine-tuning, all outputs are correct.}
\label{fig:legal_element_extraction_25}
\end{figure*}

In Figures~\ref{fig:legal_element_extraction_1} to \ref{fig:legal_element_extraction_25}, we present inference results from legal element extraction tasks, comparing the base model with the fine-tuned model. Several interesting findings emerge, with key observations as follows:

\begin{itemize}
    \item \textbf{Incomplete extraction.} As illustrated in Figure~\ref{fig:legal_element_extraction_1}, Figure~\ref{fig:legal_element_extraction_2}, and Figure~\ref{fig:legal_element_extraction_5}, the base model tends to extract incorrect or incomplete legal elements. This issue is particularly noticeable with less common or specialized legal elements such as ``armed fight'', ``confession'', and ``single defendant''. The incompleteness primarily arises from two key reasons: First, the base model lacks comprehensive and specialized legal knowledge since it has not undergone dedicated fine-tuning on legal domain datasets, making it challenging for the model to recognize and correctly identify specialized legal terms present in the context. Second, without targeted training on legal element extraction tasks, the model naturally defaults to extracting general or more obvious elements. As a result, specialized legal terminology or nuanced legal concepts, such as details related to ''smugglers'' or other professionally-specific elements, are often overlooked or misinterpreted. Addressing these challenges requires targeted domain-specific fine-tuning, enabling the model to develop a deeper understanding of complex and specialized legal elements, thereby improving extraction completeness and accuracy.

    \item \textbf{Incorrect extraction.} As illustrated in Figure~\ref{fig:legal_element_extraction_9}, the base model incorrectly extracts the element ``pronation'' as ``1'', while the ground truth label is ``8''. In contrast, our fine-tuned model successfully identifies the correct label. Similarly, in Figure~\ref{fig:legal_element_extraction_16}, the base model inaccurately extracts the legal element ``self surrender'', even though this element is not mentioned in the provided text. Again, our fine-tuned model correctly avoids this error. Furthermore, as shown in Figure~\ref{fig:legal_element_extraction_23}, the base model mistakenly identifies several non-existent elements, including ``forgiveness'', ``confession and punishment'', and ``lenient punishment'', despite these terms being absent from the text. Our fine-tuned model effectively mitigates this issue. These extraction errors likely stem from the base model's insufficient understanding and inadequate training in legal domain knowledge, highlighting the importance of specialized domain adaptation for improved accuracy.

    \item \textbf{Incorrect format conversion.} Although we explicitly emphasize in the instructions that sentencing outputs—such as fixed-term imprisonment and probation periods—should consistently be expressed in months, the base model frequently fails to adhere to this requirement. For instance, as shown in Figure~\ref{fig:legal_element_extraction_4}, the base model incorrectly outputs a fixed-term imprisonment of ``9 months'' directly as ``9'' without clearly specifying the unit, and similarly represents a probation period of ``12 months'' merely as ``12''. This issue also appears in Figure~\ref{fig:legal_element_extraction_9}, where the base model again ambiguously outputs imprisonment as ``8'' instead of clearly stating ``8 months'', and probation as ``12'' rather than explicitly as ``12 months''. Moreover, as depicted in Figure~\ref{fig:legal_element_extraction_15}, the base model mistakenly outputs ``5'' for fixed-term imprisonment, which should have been correctly converted to months as $5 \times 12 = 60$. Likewise, in Figure~\ref{fig:legal_element_extraction_23}, probation is incorrectly presented as ``1'', while the correct conversion should be $1 \times 12 = 12$ months. In contrast, our fine-tuned model effectively resolves these formatting inconsistencies, ensuring outputs adhere precisely to the instructed standards.

    \item \textbf{Unreasonable extraction.} As shown in Figure~\ref{fig:legal_element_extraction_7}, the base model exhibits unreasonable outputs by repeatedly extracting unmentioned legal elements such as ``accomplice'' and ``criminal record and misconduct'', assigning values despite their absence from the original text. Furthermore, Figure~\ref{fig:legal_element_extraction_23} illustrates another notable issue: the model generates a completely nonexistent legal element, ``no adverse effects'', which is not included in the predefined element list, indicating a clear case of hallucination. These unreasonable and erroneous outputs highlight the necessity of fine-tuning the base model on high-quality, high-confidence datasets specifically curated for legal element extraction tasks, thereby reducing such hallucinations and enhancing extraction reliability.

    \item \textbf{Unreadable output.} As demonstrated in Figure~\ref{fig:legal_element_extraction_20}, the base model occasionally produces highly disorganized and difficult-to-read outputs, significantly reducing readability and usability. Even after supervised fine-tuning (SFT), some outputs may still contain partially garbled or inconsistent formatting issues. These persistent readability problems motivate us to further employ reinforcement learning (RL) techniques, as described in the next subsection, to systematically enhance both the formatting quality and overall readability of the extracted legal elements.

\end{itemize}

To this end, we believe that constructing high-quality datasets is essential, as it enables the model to efficiently learn specialized domain knowledge and establishes a strong foundational capability, thereby supporting improved performance in subsequent, more complex tasks. In future work, we plan to further enhance model optimization by integrating reinforcement learning methods to refine output quality, reduce formatting inconsistencies, and enhance the overall reliability and interpretability of the extracted legal elements.

\clearpage
\newpage

\subsection{Reinforcement Learning}

\begin{figure*}[!b]
\centering
\includegraphics[width=0.9\linewidth]{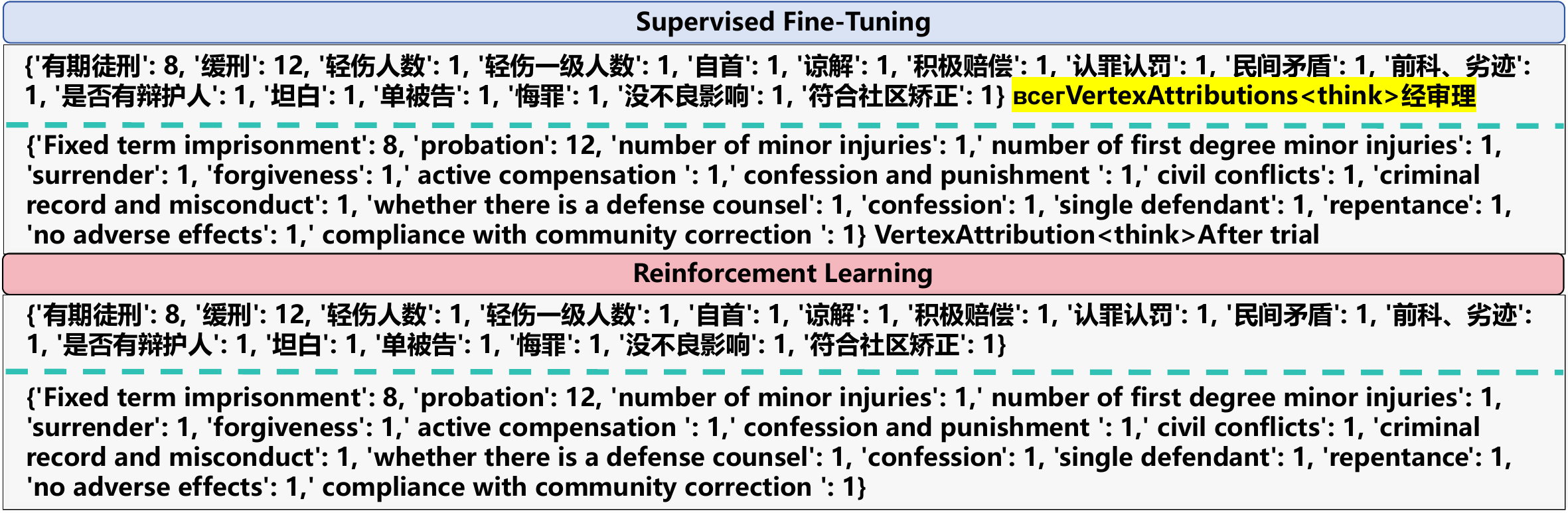}
\caption{Example of garbled output after SFT, exhibiting irregular characters that severely degrade readability.}
\label{fig:rl_format_example_1}
\vspace{-0.5em} 
\end{figure*}

\begin{figure*}[!b]
\centering
\includegraphics[width=0.9\linewidth]{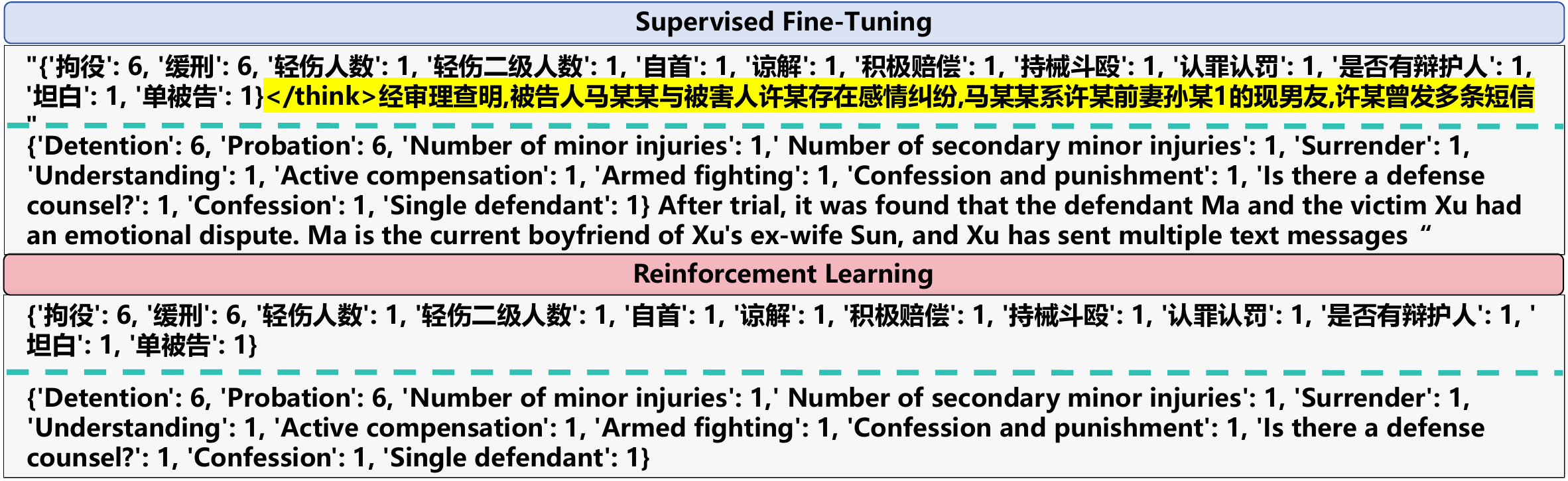}
\caption{An example of garbled output after SFT, where input information is mistakenly included in the output, complicating subsequent processing steps.}
\label{fig:rl_format_example_2}
\end{figure*}

In the legal element extraction tasks, we observe that even after supervised fine-tuning (SFT), some inference outputs still contain garbled or improperly formatted results. For instance, as shown in Figure~\ref{fig:rl_format_example_1}, the output includes unreadable characters, and in Figure~\ref{fig:rl_format_example_2}, the model mistakenly reproduces portions of the input content. These non-standard outputs severely degrade readability and significantly hinder subsequent standardization processes. To address these formatting issues, we further employ reinforcement learning (RL), specifically the GRPO algorithm~\citep{shao2024deepseekmath}, to normalize the outputs. Specifically, we design a reward function that penalizes the model when its output includes incorrect or irrelevant information such as garbled text or duplicated input content, while rewarding clear, consistent, and properly formatted outputs. After applying RL optimization, we observe that the issue of garbled or non-standard outputs has been substantially resolved, resulting in improved readability and streamlined post-processing.

\subsection{Retrieval-Based Augmentation}

\begin{figure*}[!t]
\centering
\includegraphics[width= 1.0\linewidth]{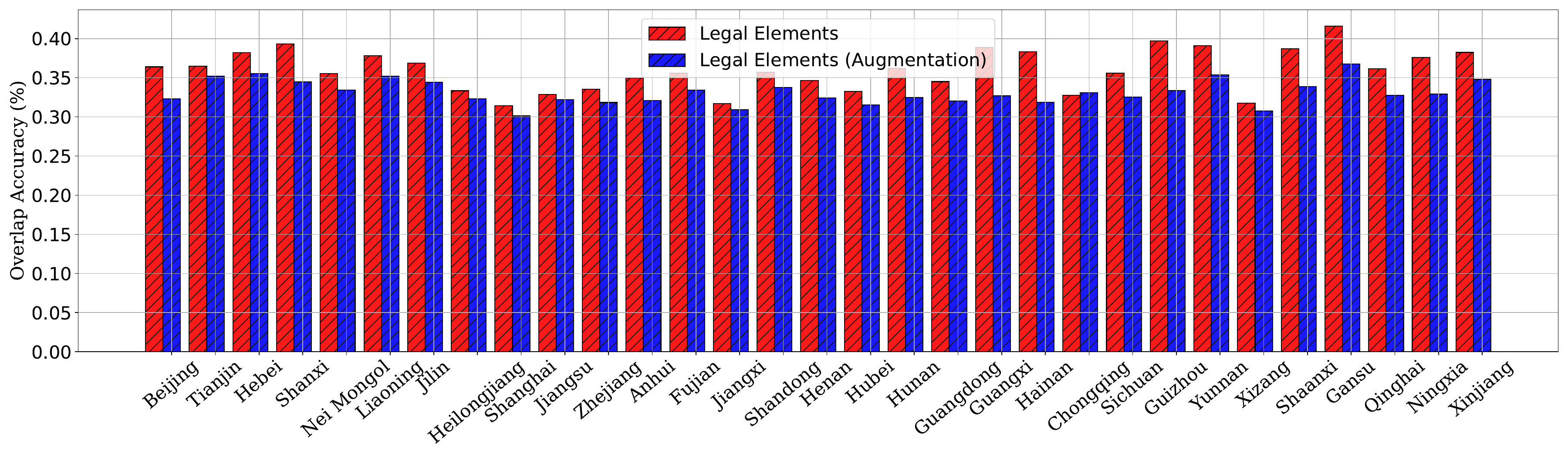}
   \caption{Comparison of retrieval results using original and augmented legal elements, highlighting the overlap accuracy with the ground truth elements.
}
\label{rag_legal_elements}
\end{figure*}

\begin{table*}[!ht]
    \centering
    \caption{Table showing examples of legal elements along with detailed augmentations and explanations. The first column lists various legal concepts, and the second column provides expanded definitions and contextual explanations in both Chinese and English. }
    \label{tab:legal_elements_examples}
    \renewcommand{\arraystretch}{1.2}  
    \resizebox{1.0\textwidth}{!}{  
        \begin{tabular}{c|>{\centering\arraybackslash}p{0.9\textwidth}}  
            \toprule
            Legal elements & Legal elements (Augmentation) \\
            \midrule 
            \begin{CJK*}{UTF8}{gkai}管制\end{CJK*} & \begin{CJK*}{UTF8}{gkai}管制（指某人被限制自由，但不完全剥夺其自由的刑罚，通常适用于罪行较轻的犯罪分子，可能包括定期报告、限制居住区域等条件）\end{CJK*} \\
            Supervision& Supervision (a sentence in which a person's freedom is restricted but not completely deprived of his or her freedom. It is usually applied to criminals with less serious crimes and may include conditions such as regular reporting and restrictions on the area where he or she lives.) \\
            \begin{CJK*}{UTF8}{gkai}拘役\end{CJK*} & \begin{CJK*}{UTF8}{gkai}拘役（指对某人进行短期的限制自由的刑罚，通常为1个月至6个月，实施拘役时，嫌疑人仍然在一定程度上保留自由，但会受到某些限制，如不得随意离开指定地点等）\end{CJK*} \\
            Detention& Detention (a short-term sentence that restricts a person's freedom, usually ranging from 1 to 6 months. During detention, the suspect still retains some freedom but faces certain restrictions, such as not being allowed to leave designated areas.) \\
            \begin{CJK*}{UTF8}{gkai}有期徒刑\end{CJK*} & \begin{CJK*}{UTF8}{gkai}有期徒刑（指法院判决某人必须在监狱服刑一定年限的刑罚，通常为几年，服刑期满后，可以重获自由，但刑期长短会根据犯罪的性质和严重程度来决定）\end{CJK*} \\
            Fixed-term imprisonment& Fixed-term imprisonment (a sentence in which the court orders a person to serve a specified term in prison, usually several years. After serving the sentence, the person may regain their freedom, with the length of the term depending on the nature and severity of the crime.) \\
            \begin{CJK*}{UTF8}{gkai}无期徒刑\end{CJK*} & \begin{CJK*}{UTF8}{gkai}无期徒刑（指法院判决某人终身在监狱服刑，虽然理论上没有刑期限制，但可以在服刑一定年限后申请假释，最终是否释放取决于罪犯表现及其他相关因素）\end{CJK*} \\
            Life imprisonment& Life imprisonment (a sentence in which the court orders a person to serve a life term in prison. Although there is theoretically no time limit for the sentence, parole can be applied for after serving a certain number of years, and the release depends on the offender's behavior and other factors.) \\
            \begin{CJK*}{UTF8}{gkai}死刑\end{CJK*} & \begin{CJK*}{UTF8}{gkai}死刑（指法院判决某人执行死刑，意味着对犯罪分子实施致命处罚，通常用于极其严重的犯罪行为，如故意杀人、恐怖活动等。死刑执行后，罪犯将无法再活跃于社会）\end{CJK*} \\
            Death penalty& Death penalty (a sentence in which the court orders the execution of a person, resulting in death. It is usually applied to extremely serious crimes such as intentional homicide or terrorism. After execution, the criminal can no longer be active in society.) \\
            \begin{CJK*}{UTF8}{gkai}缓刑\end{CJK*} & \begin{CJK*}{UTF8}{gkai}缓刑（指法院在判定某人有罪的情况下，暂时不执行刑罚，而是给予一定的观察期，如果在观察期内没有再犯，可以免于执行刑罚，但若在缓刑期内再犯，可能会被执行原定刑罚）\end{CJK*} \\
            Suspended sentence& Suspended sentence (a sentence where the court temporarily refrains from executing the punishment, giving the person a probation period. If the person does not commit any further offenses during this period, the punishment may not be executed. However, if the person reoffends, the original sentence may be enforced.) \\
            \begin{CJK*}{UTF8}{gkai}附带民事诉讼\end{CJK*} & \begin{CJK*}{UTF8}{gkai}附带民事诉讼（指在刑事案件审理过程中，受害人或其他相关方提出的民事赔偿请求，法院可以在审理刑事案件的同时，处理相关民事诉讼问题）\end{CJK*} \\
            Civil lawsuit attached & Civil lawsuit attached (a civil compensation claim filed by the victim or other relevant parties during the trial of a criminal case. The court can handle the related civil lawsuit issues while hearing the criminal case.) \\
            \begin{CJK*}{UTF8}{gkai}轻微伤人数\end{CJK*} & \begin{CJK*}{UTF8}{gkai}轻微伤人数（指在某些案件中，受害人受到的伤害较轻，但依然需要评估伤害程度和责任分配，通常由医疗鉴定机构进行评估）\end{CJK*} \\
            Minor injury count& Minor injury count (refers to cases where the victim suffers light injuries, but the degree of injury and responsibility allocation still need to be assessed, usually by a medical evaluation agency.) \\
            \begin{CJK*}{UTF8}{gkai}轻伤人数\end{CJK*} & \begin{CJK*}{UTF8}{gkai}轻伤人数（指在案件中，受害人遭受的伤害为轻度，属于刑法中规定的轻伤范畴，通常会对加害人进行一定的刑罚处罚）\end{CJK*} \\
            Light injury count& Light injury count (refers to cases where the victim suffers light injuries, falling under the category of minor injuries as defined by criminal law. Perpetrators are usually penalized accordingly.) \\
            \begin{CJK*}{UTF8}{gkai}轻伤一级人数\end{CJK*} & \begin{CJK*}{UTF8}{gkai}轻伤一级人数（指受害人在案件中遭受轻伤，伤情较为严重，但尚不构成重伤，具体伤情可根据伤残程度分级评定）\end{CJK*} \\
            Level 1 light injury count& Level 1 light injury count (refers to victims who sustain more serious light injuries, but not severe enough to constitute a major injury. The specific degree of injury can be graded based on disability.) \\
            $\cdots$ & $\cdots$ \\
            \bottomrule
        \end{tabular}
    }
\end{table*}

\begin{figure*}[!t]
\centering
\includegraphics[width= 1.0\linewidth]{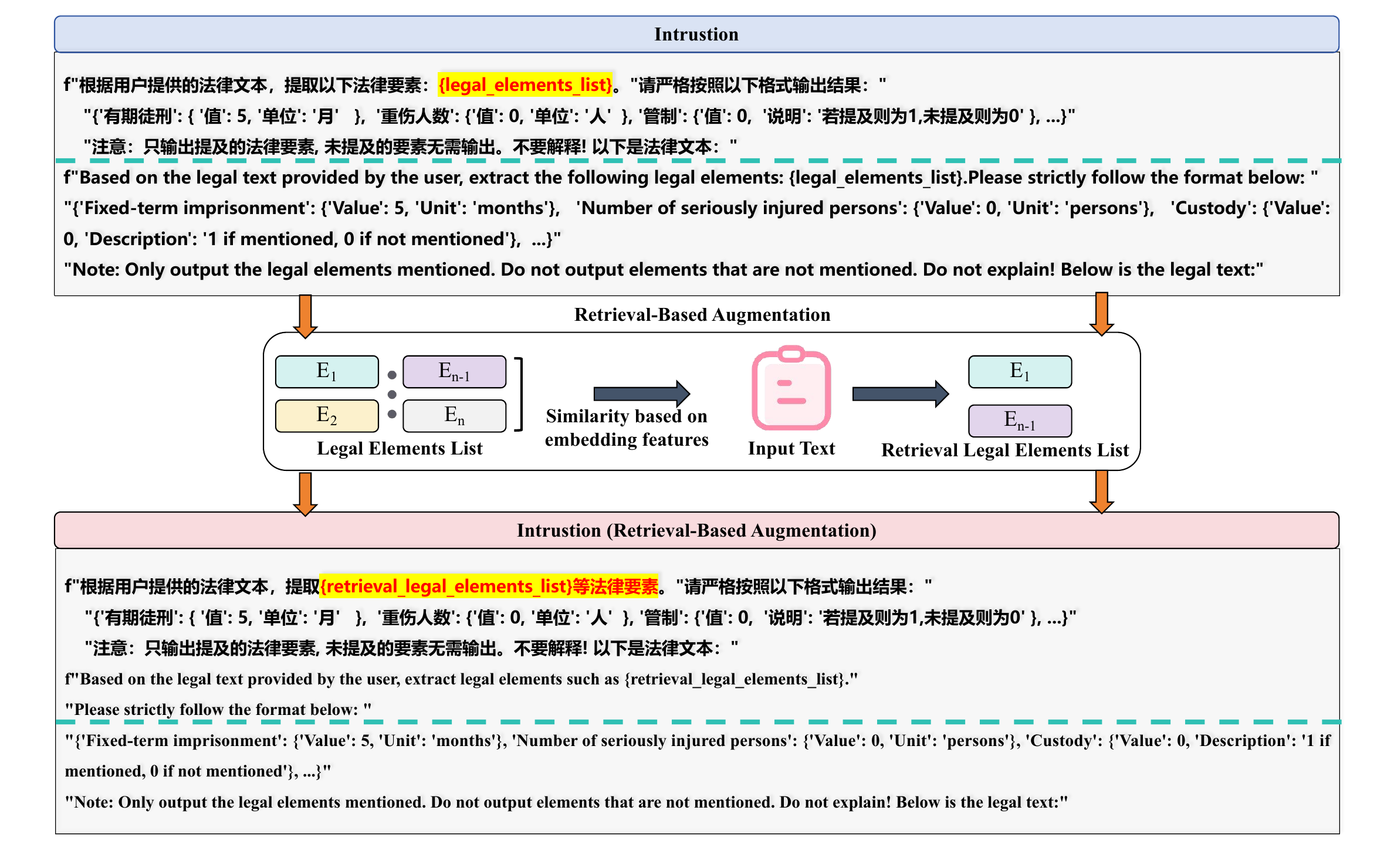}
   \caption{The overall framework of retrieval-based augmentation, where similarity is calculated based on embedding features to retrieve the subset of legal elements most relevant to the input text.
}
\label{fig:framework_rag_legal_elements}
\end{figure*}

Legal documents typically contain a vast amount of information, including basic details such as court and presiding judge information, as well as comprehensive case specifics and related facts. Although we have formatted the data (c.f. Section~\ref{sec:training_data}) to retain only the key content of the case, the resulting text is still quite lengthy, often exceeding 5,000 words. Simplifying the text further proves challenging, as legal documents vary significantly in structure and content. Removing sections based on certain keywords may inadvertently exclude crucial information, including key legal elements. Moreover, establishing universally applicable rules for different types of legal documents is complex and prone to overlooking important details. Moreover, the number of legal elements varies significantly across documents, further increasing the input token count, as illustrated in the first column of Table~\ref{tab:legal_elements_examples}. This raises a straightforward but important question: while large language models (LLMs) can accommodate longer input sequences by employing techniques such as input chunking, sliding windows, and extended positional encoding to increase the number of tokens~\citep{he2024two, zhang2025found}, the effectiveness of longer inputs is still limited by the inherent capacity of the model. Existing studies suggest that increasing the input length beyond a certain threshold may lead to degraded performance~\citep{zhang2024more}. Specifically, longer input sequences introduce significant computational complexity, which can hinder the model’s efficiency and overall performance. The overall framework is shown in Figure~\ref{fig:framework_rag_legal_elements}, where the instructions are modified based on the retrieved legal elements.

In our experiments, we observed that as the input length increases, both the inference time and the likelihood of degraded results increase exponentially. The simplest approach is to split the text into \(n\) parts, but this results in an operation time that is \(n\) times longer than the original input. Motivated by the promising performance of retrieval-augmented generation in LLMs, we propose a straightforward method to retrieve the key legal elements that match the input text, thereby augmenting efficiency. We first calculate the embeddings of each legal element using a BERT-based model~\citep{bge_embedding}. The input text is then split into $n$ parts based on the number of tokens, and the text embeddings are computed in the same way as the legal element embeddings. We do not use a semantic text filtering approach, as we find that it introduces unnecessary complexity without significantly improving performance. Finally, each text part is matched with the most relevant legal element based on cosine similarity. Based on this strategy, we hypothesize that a single word and its embedding alone may not sufficiently differentiate the meaning of a sentence~\citep{miao2024enhancing}. Building on this hypothesis, we then leverage the LLM to generate augmented versions of the words~\citep{dai2025auggpt,wang2024comprehensive}, providing detailed explanations of the meanings presented in the second column of Table~\ref{tab:legal_elements_examples}. 

To evaluate the performance of legal element embeddings and their corresponding augmented embeddings, we introduce a metric called \textit{overlap accuracy}. This metric is calculated based on the proportion of true legal elements retrieved in comparison to the total number of true elements in the input text. For instance, if there are 5 true elements and all are correctly retrieved, the overlap accuracy would be 100\%. However, if only 4 out of the 5 true elements are retrieved, the overlap accuracy would be calculated as \( \frac{4}{5} = 80\% \). The comparison result is shown in Figure~\ref{rag_legal_elements}, which reveals an interesting observation: the augmented version performs even worse than the original version. We further tested several Chinese BERT-based models, but the results remained similar. We think there are two possible reasons for this outcome: (1) the existing embedding models perform well in general domains but lack specialization in the legal domain, leading to embeddings that do not effectively distinguish between legal concepts, which results in incorrect cosine similarity matching; (2) although augmenting the embeddings from words to sentences may seem logical, it fails to accurately reflect the true structure and nuances of legal documents, leading to poor matching. To this end, we plan to fine-tune the embedding model with legal knowledge to extract more fine-grained embeddings in the next version. Additionally, a more crucial step is to construct a dedicated legal knowledge base, which will enhance the accuracy of the retrieval process in RAG.


\subsection{More Discussion}

\begin{figure*}[!t]
\centering
\includegraphics[height=8cm]{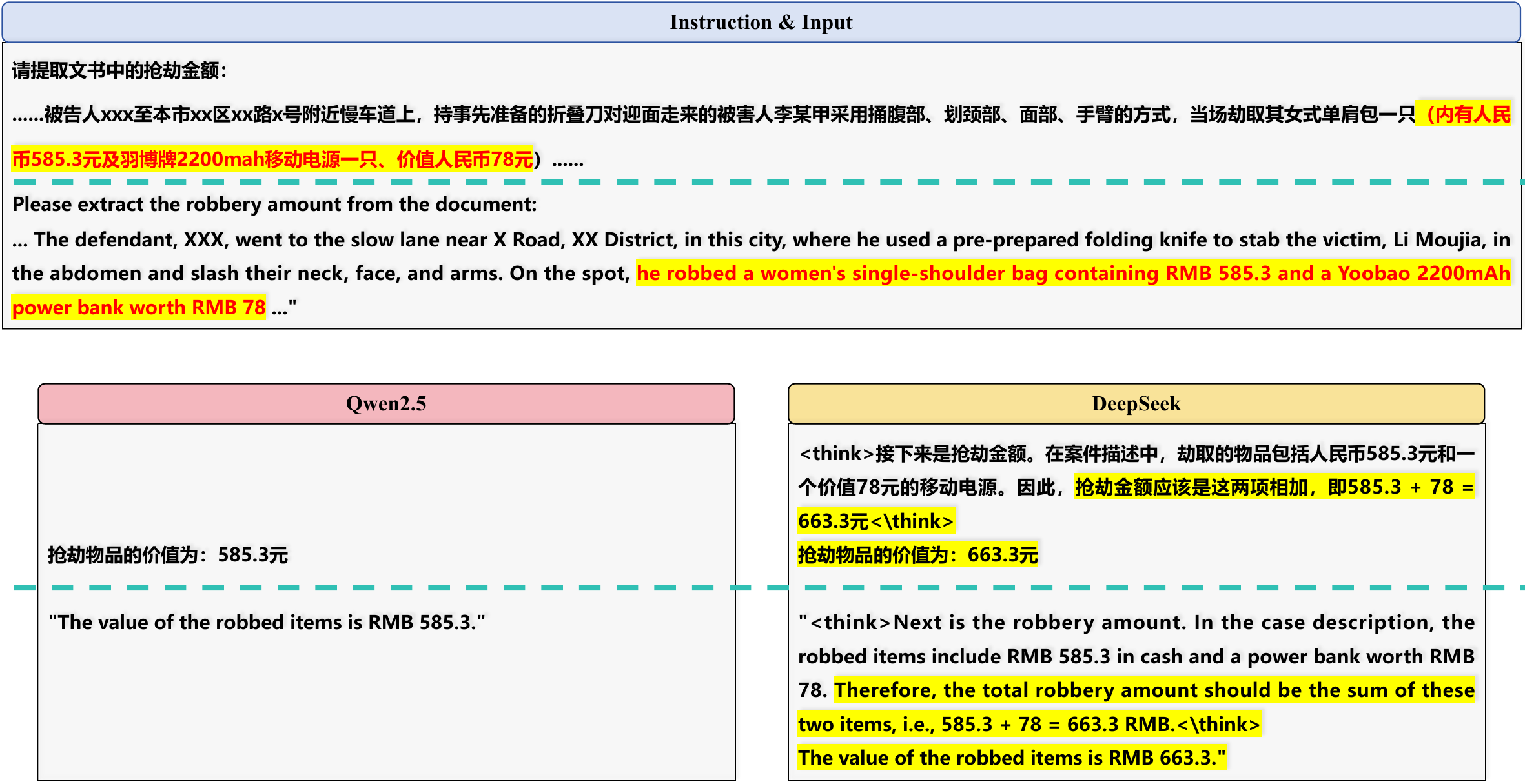}
   \caption{This example compares Qwen and DeepSeek in extracting the robbery amount from a document. The input text includes two stolen items, with a total value of 663. Qwen fails to recognize both items, extracting only the first (585.3), while DeepSeek correctly identifies both and returns the accurate sum. 
}
\label{fig:comparison_qwen_deepseek}
\end{figure*}

\begin{figure*}[!t]
\centering
\includegraphics[height=8cm]{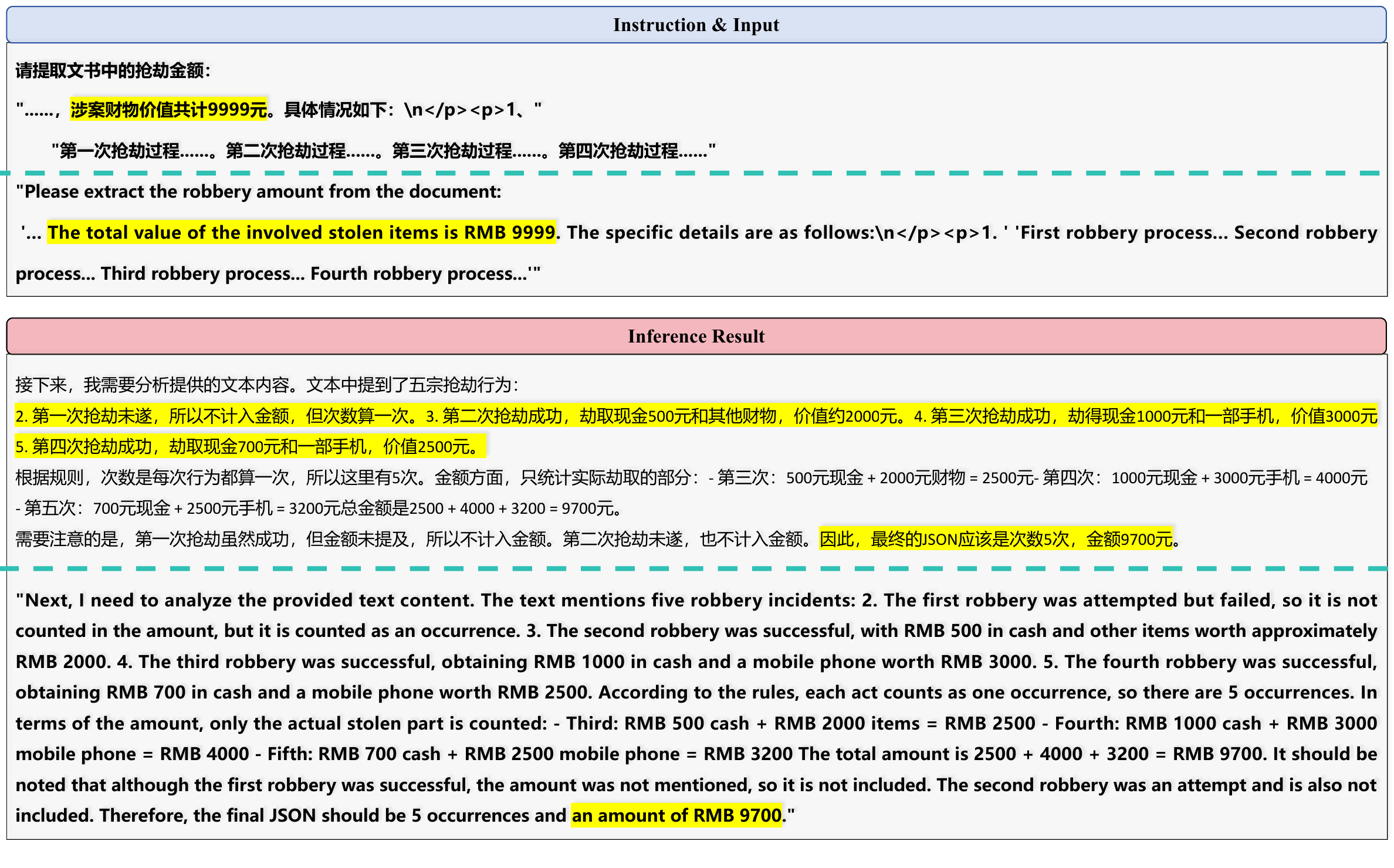}
   \caption{In this example, the document explicitly states that the total amount robbed is 9999. However, DeepSeek generates an detailed reasoning process, reconstructing the entire robbery sequence but ultimately arriving at an incorrect result due to flawed reasoning.
}
\label{fig:process_hack}
\end{figure*}

We further compare Qwen2.5-72B with DeepSeek-Distilled-14B and 70B. The evaluation involves a document related to robbery, where the task is to extract the total amount stolen. The inference results for Qwen and DeepSeek are shown in Figure~\ref{fig:comparison_qwen_deepseek}. The input text contains two stolen items, meaning the correct robbery amount should be the sum of both. However, Qwen only extracts the first item and fails to recognize the second, leading to an incorrect result. In contrast, DeepSeek successfully identifies both stolen items and correctly sums them. To further assess performance, we tested similar tasks and consistently found that DeepSeek outperformed Qwen. Based on these findings, we selected DeepSeek as the base model for fine-tuning legal knowledge.

In certain cases, we observed that DeepSeek engages in extensive reasoning processes, which do not always lead to correct results. For example, when instructed to extract the total amount robbed from a document, DeepSeek failed to identify the explicitly stated amount. Instead, it attempted to deduce the total by analyzing individual items involved in the robbery. This approach led to an incorrect calculation, as shown in Figure~\ref{fig:process_hack}. This behavior exemplifies \textit{reward hacking}, where an AI system exploits flaws or ambiguities in the reward function to achieve high rewards without genuinely solving the intended task \cite{turn0search0}. In this context, DeepSeek's reasoning process may have been influenced by an imperfect reward function during training, leading it to prioritize complex reasoning over straightforward extraction of explicitly stated information. Addressing such issues requires refining the reward function to align more closely with the desired outcomes, thereby mitigating unintended behaviors.

\section{Conclusion and Future Work}

We introduce \mymodel, our first-generation reasoning model tailored for the highly specialized Chinese legal domain. Our approach incorporates enhanced post-training techniques, including multi-stage supervised fine-tuning and reinforcement learning. To better align with real-world inspection tasks, we have carefully processed legal documents and constructed high-quality standardized training data. These efforts contribute to improved human preference alignment, legal element extraction, and structured data analysis, making \mymodel~be a promising model for instruction-following tasks. We provide \mymodel~in multiple configurations, including 14B, 32B, and 70B parameters, aiming to support a range of legal applications.

While our model demonstrates strong potential in legal research and inspection tasks, there remains room for further refinement. In future work, we plan to explore more advanced fine-tuning strategies, broaden the scope of legal data sources, and enhance the model's adaptability to complex legal reasoning scenarios. Specifically, we aim to expand the range of tasks, such as similar case recommendation, to improve practical applicability. Additionally, we plan to construct a legal knowledge base to facilitate more efficient knowledge retrieval. We hope that \mymodel~can serve as a valuable resource in legal AI development and contribute to future innovations in the field.

\clearpage
\newpage
\bibliography{colm2024_conference}
\bibliographystyle{colm2024_conference}

\end{document}